\documentclass[letterpaper,journal]{IEEEtran}
\usepackage{amsmath,amsfonts}
\usepackage{algorithmic}
\usepackage{algorithm}
\usepackage{array}
\usepackage[caption=false,font=normalsize,labelfont=sf,textfont=sf]{subfig}
\usepackage{textcomp}
\usepackage{stfloats}
\usepackage{url}
\usepackage{verbatim}
\usepackage{graphicx}
\usepackage{cite}
\usepackage{indentfirst}
\usepackage{multirow}
\usepackage{booktabs} 
\usepackage[hidelinks]{hyperref}
\usepackage{cleveref}
\usepackage{graphicx}   
\usepackage{epstopdf}   

\usepackage{xcolor}

\newcommand{\rev}[1]{#1}

\crefname{figure}{Fig.}{Figs.}
\Crefname{figure}{Fig.}{Figs.}
\IEEEoverridecommandlockouts

\begin{document}


\title{IGASA: Integrated Geometry-Aware and Skip-Attention Modules for Enhanced Point Cloud Registration}

\author{Dongxu Zhang, 
        Jihua Zhu,~\IEEEmembership{Senior Member,~IEEE,} %
        Shiqi Li,~\IEEEmembership{Student Member,~IEEE,} %
        Wenbiao Yan, %
        Haoran Xu, %
        Peilin Fan %
        and~Huimin Lu,~\IEEEmembership{Senior Member,~IEEE}
}

\markboth{IEEE Transactions On Circuits and Systems For Video Technology}%
{Shell \MakeLowercase{\textit{et al.}}: A Sample Article Using IEEEtran.cls for IEEE Journals}


\maketitle

\begin{abstract}

Point cloud registration (PCR) is a fundamental task in 3D vision and provides essential support for applications such as autonomous driving, robotics, and environmental modeling. Despite its widespread use, existing methods often fail when facing real-world challenges like heavy noise, significant occlusions, and large-scale transformations. These limitations frequently result in compromised registration accuracy and insufficient robustness in complex environments.
\rev{In this paper, we propose IGASA as a novel registration framework constructed upon a Hierarchical Pyramid Architecture (HPA) designed for robust multi-scale feature extraction and fusion. The framework integrates two pivotal components consisting of the Hierarchical Cross-Layer Attention (HCLA) module and the Iterative Geometry-Aware Refinement (IGAR) module. The HCLA module utilizes skip attention mechanisms to align multi-resolution features and enhance local geometric consistency. Simultaneously, the IGAR module is designed for the fine matching phase by leveraging reliable correspondences established during coarse matching. This synergistic integration within the architecture allows IGASA to adapt effectively to diverse point cloud structures and intricate transformations.}
We evaluate the performance of IGASA on four widely recognized benchmark datasets including 3D(Lo)Match, KITTI, and nuScenes. Our extensive experiments consistently demonstrate that IGASA significantly surpasses state-of-the-art methods and achieves notable improvements in registration accuracy. This work provides a robust foundation for advancing point cloud registration techniques while offering valuable insights for practical 3D vision applications. The code for IGASA is available in \href{https://github.com/DongXu-Zhang/IGASA}{https://github.com/DongXu-Zhang/IGASA}.

\end{abstract}

\section{Introduction}

Point Cloud Registration (PCR) is a critical task in 3D computer vision, aimed at aligning point clouds captured from different viewpoints or at different time instances to achieve data fusion and complementary information integration~\cite{zhang2024comprehensive,ZHANG2026133318}. This process plays a critical role in various applications, such as autonomous driving~\cite{wang2022residual,huang2024kdd}, robot navigation~\cite{wang2022efficient,wang2021deep}, and environmental modeling, where accurate alignment is essential for high-precision perception and decision-making. However, the inherent challenges associated with point cloud data-such as noise, occlusion, non-uniform sampling, and complex transformations ($e.g.,$ large rotations and scaling), significantly hinder both the accuracy and efficiency of registration methods~\cite{bai2021pointdsc}~\cite{qin2022geometric,she2024pointdifformer},~\cite{yew2022regtr}. These issues complicate the registration process, leading to potential inaccuracies and inefficiencies.

\IEEEpubidadjcol

Traditional registration techniques, notably the Iterative Closest Point (ICP) algorithm and its variants~\cite{jost2002fast,vizzo2023kiss},~\cite{sharp2002icp} rely on minimizing point-to-point distances. Although ICP is widely used, it exhibits significant sensitivity to the initialization of parameters. This sensitivity can lead to convergence on local minima, especially when the initial guess is poor or when the data is heavily corrupted by noise and occlusion. Variants of ICP attempt to mitigate these issues, but they still fall short under severe conditions~\cite{yang2015go}.

\rev{Rigid point cloud registration aims to determine the optimal rotation and translation to align two overlapping point clouds. Correspondence based methods\cite{qin2022geometric,xu2024igreg} generally tackle this task by establishing reliable point matches through a multi stage pipeline. The process typically begins with detecting keypoints and encoding their local geometric properties into feature descriptors. These descriptors are subsequently matched via nearest neighbor search to generate initial correspondences. The final transformation is then computed using robust estimators, such as RANSAC\cite{fischler1981random}, to ensure accurate alignment despite potential outliers.}


\begin{figure}[t]
    \centering
    \includegraphics[width=0.85\linewidth]{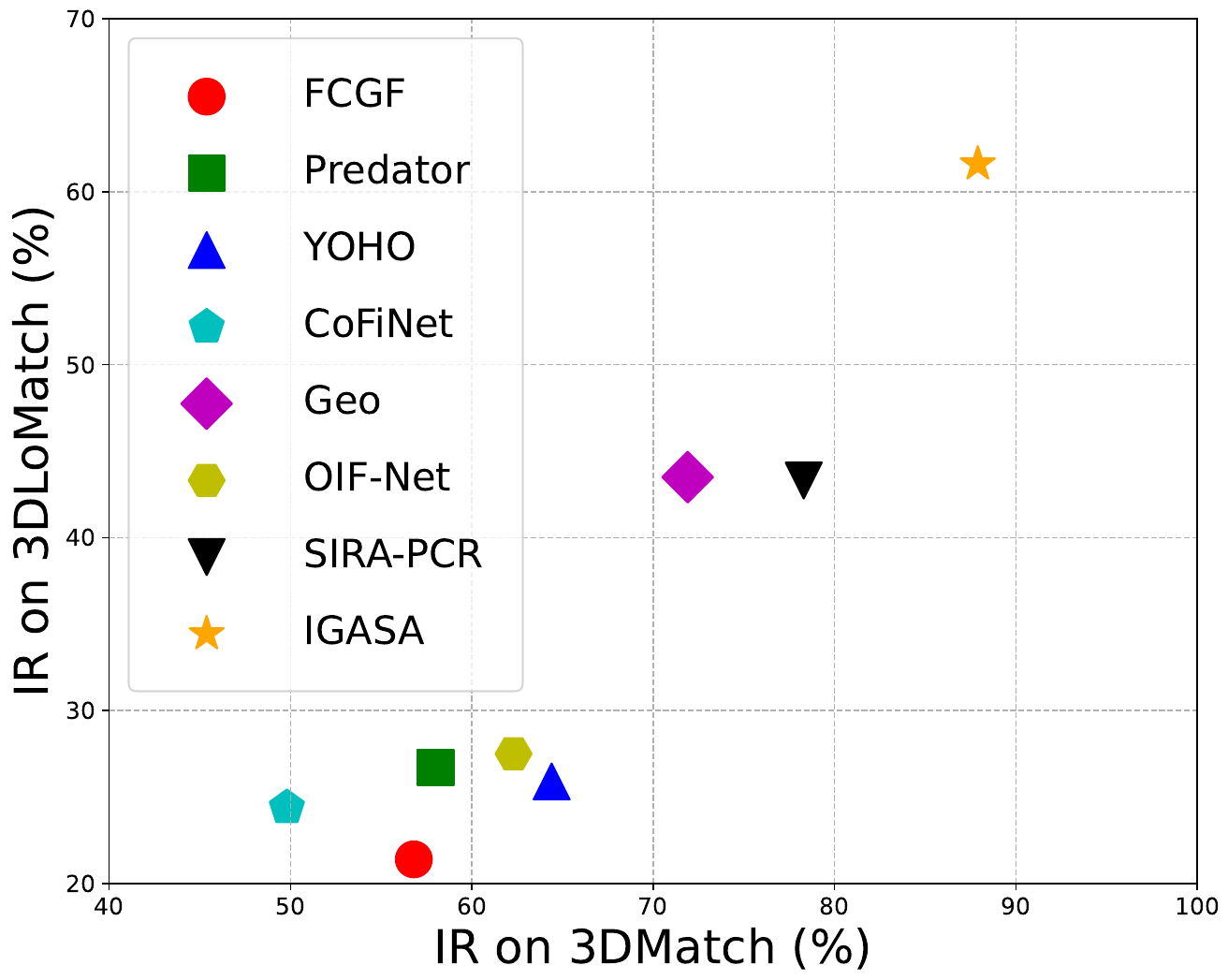}
    \caption{The Inlier Ratio (IR) is plotted on the x-axis for 3DMatch and on the y-axis for 3DLoMatch. IGASA stands out by consistently achieving the highest IR.}
    \label{figurelabel}
    \vspace{-5mm}
\end{figure}


\rev{Recently, deep learning methods have emerged as a promising alternative, demonstrating powerful capabilities in adaptive feature extraction~\cite{zhang2024adaptive}, and offering end-to-end solutions that learn feature representations directly from raw point cloud data~\cite{huang2021predator,yu2021cofinet}.}
Convolutional Neural Networks (CNNs) have been successfully applied for local feature extraction~\cite{9010002}. However, CNNs are constrained by fixed receptive fields that limit their ability to capture long-range dependencies. In contrast, transformer-based architectures ~\cite{qin2022geometric,yang2022one,yu2023rotation},\cite{chen2023sira,huang2017coarse} have demonstrated efficacy in capturing global context, a capability that is particularly advantageous for modeling the overall shape and structure of point clouds. Moreover, advances in deformable image registration have illustrated the benefits of integrating hierarchical and attention mechanisms, where short range self-attentions are complemented by cross-attention strategies to capture broader contextual information~\cite{xie2023cross,xu2024igreg},~\cite{ren2022corri2p,an2024ol}. These methods underscore the necessity of incorporating multi-scale representations to accurately model deformation fields and, by extension, improve point cloud registration.

Motivated by these developments, we propose IGASA, builds on these advancements by integrating a multi-scale hierarchical pyramid architecture (HPA) with two key modules: the Hierarchical Cross-Layer Attention (HCLA) module and the Iterative Geometry-Aware Refinement (IGAR) module. 

The HPA is designed to extract and fuse features at multiple scales, ensuring that both global context and fine local details are effectively captured. Within this framework, the HCLA module employs a skip attention mechanism to seamlessly align multi-resolution features, thereby enhancing local geometric consistency across scales. In practice, the skip attention mechanism weights the contributions from various layers based on their relevance to the registration task, thereby reducing the negative impact of noise and occlusion. 

Complementarily, the IGAR module is specifically designed for the fine matching phase. In each iteration, \rev{the module utilizes geometric cues such as spatial distribution and geometric alignment consistency to assess and enhance the reliability of correspondences. By employing an alternating optimization strategy, the module progressively updates the rotation and translation parameters.} This design enables IGASA to adapt to diverse point cloud structures and complex transformations, significantly improving registration accuracy while maintaining computational efficiency.

Our method's efficacy is rigorously validated across three widely recognized benchmark datasets: 3DMatch~\cite{zeng20173dmatch},3DLoMatch~\cite{huang2021predator}, KITTI~\cite{geiger2013vision}, and nuScenes~\cite{caesar2020nuscenes}. Experimental results consistently demonstrate that our method substantially outperforms state-of-the-art methods, achieving superior registration accuracy and robustness even under challenging conditions. As illustrated in~\cref{figurelabel}, IGASA achieves the highest Inlier Ratio across various overlapping scenes, underscoring its robust performance.

Beyond accuracy, the computational efficiency of IGASA is a key aspect of its design. The framework is engineered to strike an optimal balance between complexity and performance, making it well-suited for real-time applications. IGASA is optimized for rapid convergence within a minimal number of iterations, ensuring overall efficiency without compromising its high accuracy.

Our main contributions are summarized as follows:
\begin{itemize}

\item We introduce the HCLA module, a novel component engineered to leverage skip attention for precisely aligning multi resolution features, ensuring both local and global geometric consistency across point cloud structures.

\item \rev{We propose the IGAR module, an iterative refinement strategy that utilizes spatial geometric consistency. By employing an alternating optimization method, it actively suppresses outliers and significantly enhances the precision of the final pose estimation.}

\item \rev{We design an innovative HPA framework that seamlessly integrates efficient multi-scale feature extraction with robust registration capabilities, making the model exceptionally well suited for complex real world scenarios.}

\item \rev{We conduct extensive evaluations on widely recognized benchmarks, including 3D(Lo)Match, KITTI, and nuScenes. Experimental results demonstrate that our proposed IGASA significantly outperforms state of the art methods, achieving superior accuracy and robustness.}

\end{itemize}

\section{RELATED WORK}
\label{sec:formatting}

\subsection{Point Cloud Registration}

\rev{PCR is pivotal in 3D computer vision, underpinning applications such as autonomous driving and SLAM~\cite{bai2021pointdsc, jiang2024gtinet}. Traditionally, optimization based methods, exemplified by the ICP algorithm~\cite{sharp2002icp} and its variants~\cite{jost2002fast, yang2015go}, rely on nearest neighbor search to iteratively minimize alignment errors. However, the reliance on spatial proximity renders the optimization objective inherently non convex. Consequently, these methods are highly susceptible to converging to local minima when confronted with large initial misalignments or sparse data~\cite{vizzo2023kiss}, necessitating precise initialization to guarantee convergence.}

\rev{Recent advancements in deep learning have shifted focus toward end-to-end learning frameworks. While early methods utilized local descriptors~\cite{gojcic2019perfect,choy2019fully,sun2025hyperpoint,sun2026alignadaptrethinkingparameterefficient,han2025rethinking}, state-of-the-art methods increasingly leverage Transformer architectures to model long-range dependencies and global context~\cite{qin2022geometric,lu2019deepvcp,zhang2026pointcot,sun6064487curve3d}. Despite their proficiency in handling low-overlap scenarios, these deep architectures often encounter a critical limitation known as the semantic gap. As network depth increases to capture high-level semantics, fine-grained geometric details are frequently diluted due to aggressive downsampling operations. This loss of high-frequency cues complicates precise local alignment, particularly in complex scenes with varying point densities.}

\rev{To address these challenges, we propose the IGASA framework. Distinct from prior arts, our method bridges the semantic gap via a HCLA module that explicitly preserves multi-scale features. Furthermore, we integrate an IGAR module to dynamically enforce geometric consistency, ensuring robust and high-precision registration.}

\begin{figure*}[t]
  \centering
  \includegraphics[width=1\linewidth]{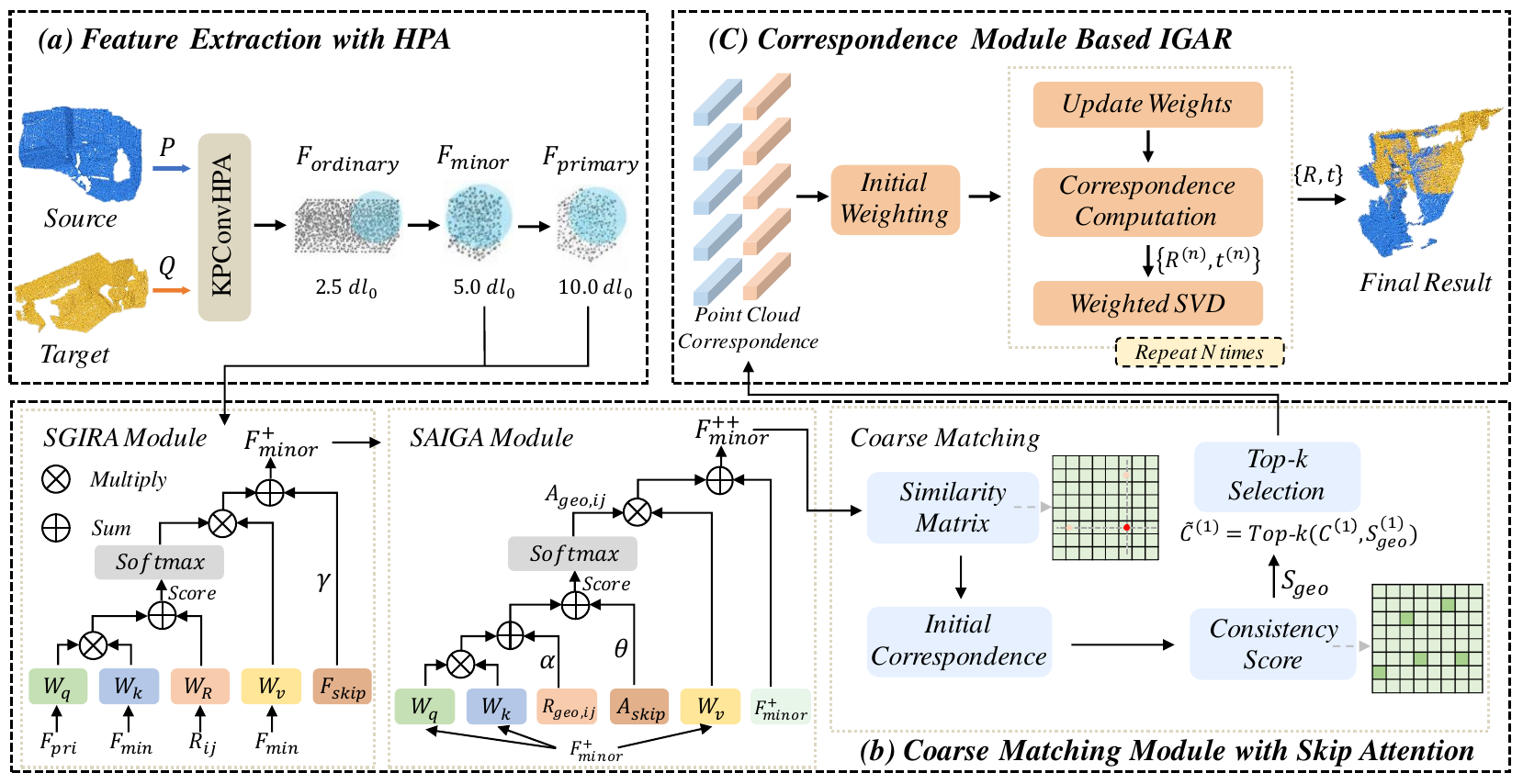}
  \caption{\rev{The overview of the proposed IGASA framework. The pipeline operates in three stages. \textbf{(a)}, the HPA module processes the source $P$ and target $Q$ inputs to construct a hierarchical feature pyramid ($F_{ordinary}, F_{minor}, F_{primary}$) with progressively expanding receptive fields. \textbf{(b)}, the HCLA module performs coarse matching by utilizing the SGIRA to fuse global semantics with local geometry, followed by the SAIGA for feature refinement. \textbf{(c)}, the IGAR module executes the fine registration phase, where an iterative optimization loop (repeated $N$ times) dynamically updates correspondence weights to suppress outliers and estimate the final transformation $\{R, t\}$.}}
  \label{figurelabe2}
  \vspace{-5mm}
\end{figure*}

\subsection{Transformer and Skip-Attention Mechanism}

\rev{Transformer-based architectures have significantly advanced point cloud registration by leveraging self-attention to capture long-range dependencies and global context~\cite{qin2022geometric, yew2022regtr, xu2024igreg}. These models~\cite{chen2025uni,lan2026performance,lan2026reco,xue2024integrating} demonstrate robustness in handling complex spatial variations and noise. However, standard self-attention mechanisms primarily focus on intra-layer interactions, often neglecting the effective integration of multi-resolution features required for fine-grained alignment.}

\rev{Although traditional skip connections~\cite{agarwal2023attention, wen2020point},~\cite{wang2024neighborhood, yuan2023egst} attempt to mitigate this by transferring high-resolution features from shallow layers, they typically rely on naive fusion strategies such as concatenation or summation~\cite{chen2019multi, horache20213d}. These methods suffer from a critical semantic gap, where the discrepancy between low-level geometric cues ($e.g.,$ edges, local density) and high-level semantic embeddings leads to suboptimal feature fusion. Consequently, critical geometric details are often diluted during the fusion process, limiting the model's precision~\cite{wang2022improving, yuan2023pointmbf}.}

\rev{To bridge this gap, we introduce a skip-attention mechanism inspired by multi-scale fusion strategies~\cite{lin2022ds, zhou2022sewer}. Unlike direct feature transfer, our method utilizes global semantic information derived from deep layers to dynamically modulate high-resolution features within a pyramid architecture. By computing cross-layer attention weights, the module selectively emphasizes geometrically relevant regions while suppressing noise and irrelevant background.}

\subsection{\rev{Coarse-to-Fine Refinement}}

\rev{Coarse-to-Fine strategies have become the paradigm for robust registration, leveraging pyramid architectures~\cite{ghiasi2019fpn},~\cite{nie2022pyramid,gong2025med} to resolve scale ambiguity. Particularly, recent frameworks like PYRF-PCR~\cite{zhang2023pyrf} have validated the efficacy of this strategy in complex outdoor environments by progressively refining poses from low to high resolution.}

\rev{Existing methods utilize these hierarchical designs to establish correspondences~\cite{zhang2025ascot,zhang2026chain,zhang2026notallqueries}. For instance, methods like CoFiNet~\cite{yu2021cofinet} and KPConv-based architectures~\cite{thomas2019kpconv,huang2022imfnet} use multi-scale fusion to ensure the capture of both local geometry and global context. Similarly, Transformer-based methods such as GeoTransformer~\cite{qin2022geometric}, EGST~\cite{she2024pointdifformer}, and RegTR~\cite{yew2022regtr} leverage hierarchical dependencies to bolster coarse matching accuracy. However, the subsequent refinement stage in these methods often relies on RANSAC or hard-thresholding techniques to filter outliers. These explicit rejection strategies can be computationally expensive and sensitive to noise, often discarding valid correspondences in low-overlap regions.}

\rev{In contrast to these hard-decision methods, we propose the IGAR module. Instead of relying on noisy normal estimation or hard pruning, IGAR employs a dynamic geometric consistency weighting strategy. This acts as a soft suppression mechanism that iteratively down-weights inconsistent pairs, providing a more robust solution to outlier rejection in challenging, non-uniform environments.}

\section{METHODS}

\subsection{Overview}
\rev{In this section, we introduce the proposed IGASA framework. Our method are illustrated in the \cref{figurelabe2}, the proposed IGASA framework aligns the source point cloud $P$ and target $Q$ through a three-stage pipeline. Initially, the HPA module constructs a multi-scale representation by leveraging Kernel Point Convolution (KPConv) to extract features at $F_{primary}$, $F_{minor}$, and $F_{ordinary}$. Subsequently, the HCLA module integrates the extracted features using a cross-layer attention mechanism. This process utilizes global semantics from $F_{primary}$ to strategically guide the refinement of $F_{minor}$, outputting geometry optimized features $F_{minor}^{++}$ that are used to establish robust coarse correspondences. Finally, these preliminary matches are forwarded to the IGAR module, which performs fine-grained registration. This hierarchical design ensures a seamless data flow: HPA provides the multi-scale basis, HCLA aligns these features semantically, and IGAR refines the geometric alignment, forming a coarse-to-fine registration system.}

\subsection{Hierarchical Pyramid Architecture Module}
\rev{The proposed HPA is designed to efficiently extract multi-scale features from point cloud data, enhancing registration performance by integrating fine-grained local details. HPA constructs a multi-resolution representation through a progressive encoder framework, leveraging KPConv~\cite{thomas2019kpconv} in handling irregular point distributions.
The core mechanism of HPA adapts the formulation of KPConv which defines a set of $K$ learnable kernel points $\{\tilde{x}_k | k < K\}$ constrained within a sphere of radius $R_{in}$. For an arbitrary query point $p_i$ and its supporting neighborhood $N(p_i)$, the convolution weights are determined by the geometric correlation between the neighbors and the fixed kernel points. Formally, the feature aggregation $f(p_i)$ is computed as:
\begin{equation}
f(p_i) = \sum_{p_j \in N(p_i)} W(p_j - p_i) \cdot f(p_j),
\end{equation}
where $W(\cdot)$ denotes the kernel function defined by the linear correlation $max(0, 1 - \frac{\| (p_j - p_i) - \tilde{x}_k \|}{\sigma})$. The influence distance $\sigma$ and the kernel radius $R_{in}$ serve as the critical hyperparameters governing the receptive field, which we dynamically adjust to facilitate multi-scale feature extraction.}

\rev{To construct the feature pyramid, the HPA module employs a three-stage hierarchy comprising the ordinary, minor, and primary levels. We utilize a grid subsampling strategy where the voxel size, denoted as $dl$, is progressively increased to reduce point density while expanding the semantic scope. At the initial ordinary level, the input point cloud $P$ is processed with a base voxel size of $dl_0$. In this high-resolution layer, the KPConv influence radius is set to $R_{in} = 2.5 \cdot dl_0$, a configuration for capturing fine-grained local geometry and preserving detailed spatial relationships. Subsequently, the network transitions to the minor level, where grid subsampling is applied with a voxel size of $2 \cdot dl_0$, producing features that encapsulate semi-global structures while reducing computational redundancy.
Finally, at the primary level, the point cloud is downsampled to a coarse resolution of $4 \cdot dl_0$, with the convolution radius extended to $10.0 \cdot dl_0$. The resulting multi-scale feature set is expressed as:
\begin{equation}
    F_{\mathrm{multi}} = \left\{ F_{\mathrm{ordinary}}, F_{\mathrm{minor}}, F_{\mathrm{primary}} \right\}.
\end{equation}}

\rev{By systematically scaling the convolution radius commensurate with the sampling density, the HPA ensures that the network transition its receptive field from local geometric fidelity to global semantic coherence, yielding a hierarchical feature space suitable for the subsequent attention based alignment.}

\subsection{Hierarchical Cross-Layer Attention Module}

\rev{The HCLA module is designed as the core integration engine to bridge the semantic gap between multi-resolution features. While standard concatenation fuses features mechanically, it often fails to address the misalignment between global context and local geometry. To address this, HCLA introduces a structured mechanism composed of two specialized components: Skip-Guided Inter-Resolution Attention (SGIRA) and Skip-Augmented Intrinsic Geometric Attention (SAIGA).}

\rev{SGIRA acts as a semantic filter. It leverages the deep, globally-aware features from the primary layer as a guide to weigh the high-resolution features in the minor layer. This process ensures that the model focuses only on semantically relevant local details while suppressing ambiguous background noise caused by resolution differences. Following this cross-layer filtering, SAIGA functions as a geometric sharpener. It performs self-attention on the filtered features to reinforce their intrinsic spatial distinctiveness, ensuring that the final descriptors are robust to viewpoint changes.}

\rev{After alignment, HCLA applies a geometric consistency evaluation to filter coarse matches. The initial point pairs generated from low-resolution features are refined through hypothesized rigid transformations. Subsequently, a compatibility graph is constructed to assess geometric consistency by quantifying the error between pairs, and a $top$-$k$ selection strategy is employed to retain the most reliable point pairs, ensuring accurate correspondences for subsequent fine registration.}

\subsubsection{Skip-Guided Inter-Resolution Attention}

The SGIRA module is designed to facilitate the alignment of semantics and geometry between low and high resolution features using a cross-layer attention mechanism. The feature $F_{primary}$, representing global semantics, guides the fusion of high-resolution features $F_{minor}$, which capture detailed local geometry. 

\begin{figure}[t]
  \centering
  \includegraphics[width=1\linewidth]{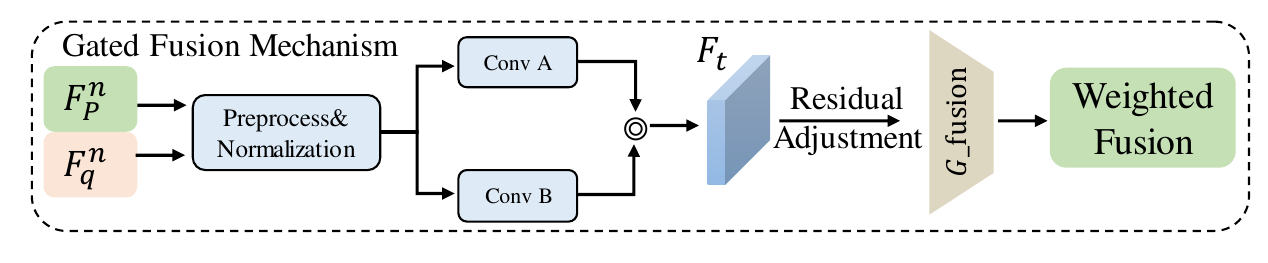}
  \caption{Details of Gated Fusion Mechanism. $F_p^n$ and $F_q^n$ are preprocessed, passed through parallel convolutional layers, combined to form $F_t$, and then refined by a residual adjustment to produce $G_{\text{fusion}}$ for weighted fusion.}
  \label{figurelabe3}
  \vspace{-5mm}
\end{figure}

\rev{To enhance this fusion, skip features from intermediate layers are incorporated to refine the distribution. While \cref{figurelabe2} (b) illustrates the logical flow of this feature integration, the specific mathematical implementation is detailed in \cref{figurelabe3}. We employ a Gated Fusion Mechanism at this interaction point, which utilizes parallel convolutional branches to dynamically weigh and combine features from different resolutions, ensuring that complementary information is effectively integrated.}
This mechanism first normalizes the input features to standardize their scale and distribution, improving the stability of subsequent operations. It then uses two convolutional branches to transform and gate the features. These branches are subsequently merged via a gating operation that computes adaptive weights, ensuring that the features from both resolutions are balanced appropriately. A residual adjustment step further enhances the features by correcting and refining them. Finally, a weighted fusion process combines these refined features, giving the final output a balanced and enriched representation.

The core of SGIRA is its attention mechanism, which not only measures semantic similarity between low and high resolution features but also incorporates geometric consistency and skip-feature guidance. The calculation process is as follows: 
\begin{equation}
    S_{ij} = \frac{Q_i K_j^T}{\sqrt{d_a}},
\end{equation}
where $S_{ij}$ reflects how closely point $i$ aligns semantically with point $j$ and $\sqrt{d_a}$ is the scaling factor. To account for geometric alignment, we define a compensation term $R_{ij}$ to reflect spatial alignment. The equation is formulated:
$R_{ij} = -\frac{\| P_i - M_j \|^2}{\sigma^2}$,
where $P_i,M_j$ represent the coordinates of the two points in the three-dimensional space. 
In addition to this, a crucial step is the weighting of the distances through the skip mechanism. Following this, we can calculate the residual error through the use of the $A_{ij}$ coefficient and the Softmax function, which normalizes the computation by applying a weight to the spatial differences.
Subsequently, we define the total residual function for optimization as:
$F_{\text{minor}}^{+} = \sum_{i=1}^{N_p} A_{ij} \cdot V_i$,
where $F_{\text{minor}}^{+}$ represents the low-frequency residual values influenced by the spatial distribution model. The residual function is given by:
\begin{equation}
F^{++}_{\text{minor}} 
= F^{+}_{\text{minor}}
+ \gamma \cdot \mathrm{SkipResidual}\!\Bigl(F^{+}_{\text{minor}},\,F_{\text{skip}}\Bigr),
\end{equation}
the term $\gamma$ scales the contribution of the skip residual component and $F_{\text{skip}}$ obtained directly from the HPA module through several skip connections.

Overall, the SGIRA module efficiently combines semantic, geometric, and skip-feature information, ensuring effective multi-scale feature fusion while maintaining computational efficiency and spatial consistency.

\subsubsection{Skip-Augmented Intrinsic Geometric Attention}

This module aims to optimize the geometric alignment within high-resolution point cloud features. SAIGA centers on refining local geometric structures. 
Starting with the feature $F^{+}_{\text{minor}}$, SAIGA performs feature projection, neighborhood alignment weight computation, and feature updates to iteratively improve feature alignment. After completing the feature projection step, the SAIGA module employs a self-attention mechanism that dynamically calculates pairwise weight distributions. The calculation of attention weights involves two key components: semantic similarity measurement and geometric distance compensation. Semantic similarity focuses on the interaction between the query vector $Q_{geo}$ and the key vector $K_{geo}$. By performing a scaled dot-product calculation, the module quantifies the correlation between each high-resolution point and its neighborhood points, as expressed by the formula:
\begin{equation}
S_{\text{geo}, ij} = \frac{Q_{\text{geo}, i} K_{\text{geo}, j}^T}{\sqrt{d_a}},
\end{equation}
where $\sqrt{d_a}$ represents the dimensionality of the query and key vectors, ensuring numerical stability during the computation. And the geometric distance weight models the contribution of the Euclidean distance between points $i$ and $j$ in the similarity computation, formulated as:
$R_{\text{geo}, ij} = -\alpha \| M_i - M_j \|^2$,
where $M_i,M_j$ denote the 3D coordinates of points $i$ and $j$, and $\alpha$ is a learnable hyperparameter that controls the influence of geometric distance.

By integrating the semantic similarity $S_{\text{geo}, ij}$ with the geometric distance compensation $R_{geo,ij}$, the SAIGA module calculates a comprehensive attention weight. Additionally, the combined attention weight is then normalized using the Softmax function, as described by:
$A_{\text{geo}, ij} = \text{Softmax}\left(S_{\text{geo}, ij} + R_{\text{geo}, ij} + \theta \cdot A_{\text{skip}}\right)$. 

Once the attention weights are computed, SAIGA performs a weighted feature aggregation over the value matrix $V_{geo}$ to update the high-resolution features. The final geometry-optimized feature representation is expressed as:
\begin{equation}
F^{++}_{\text{minor},i} 
= F^{+}_{\text{minor},i}
+ \sum_{j=1}^{N_m} A_{\text{geo}, ij} \cdot V_{\text{geo}, j},
\end{equation}
where $F^{++}_{\text{minor},i}$ represents the geometry-optimized feature of the $i-th$ high-resolution point. $F^{++}_{\text{minor},i}$ composed of the updated features for all points, serves as the final output of this step. The use of skip-augmented attention allows the model to preserve long-range dependencies while reinforcing local feature consistency.

Finally, by leveraging this attention-based feature refinement, the SAIGA module effectively yields high-resolution point representations.

\subsubsection{Superpoint matching module}

In the establishment of coarse matching relationships, we leverage the low-resolution features output to construct preliminary correspondences across the global scope. The feature representations of the source and target point clouds are denoted as $F_{\text{src}}^{(1)} = \bigl\{\, f_{\text{src},j}^{(1)} \in \mathbb{R}^d\bigr\}$ and $F_{\text{tar}}^{(1)} = \bigl\{\, f_{\text{tar},j}^{(1)} \in \mathbb{R}^d\bigr\}$, respectively. To establish these initial correspondences, for any given source point $p^{(1)}_{src,j}$, we employ a nearest neighbor search to identify the target point $p^{(1)}_{tar,k}$. \rev{The similarity between the features of the two points is defined by the Euclidean distance:
$\text{Sim}(f_{\text{src}, j}^{(1)}, f_{\text{tar}, k}^{(1)}) = \| f_{\text{src}, j}^{(1)} - f_{\text{tar}, k}^{(1)} \|_2$.
Consequently, the target point with the minimum distance is selected as the match, formalized as $p_{\text{src}, j}^{(1)} \leftrightarrow p_{\text{tar}, k}^{(1)}$,where $k = \arg\min_{k} \bigl\| f_{\text{src}, j}^{(1)} - f_{\text{tar}, k}^{(1)} \bigr\|_2$.}



Through this process, a set of preliminary matching point pairs is generated, expressed as:
$C^{(1)} = \left\{ p_{\text{src}, j}^{(1)}, p_{\text{tar}, k}^{(1)} \right\} \quad j = 1, 2, 3, \dots, N_{\text{src}}$, 
where $N_{\text{src}}$ denotes the total number of points in the source point cloud. We use a $top$-$k$ selection strategy based on a geometric consistency score to filter out erroneous matches from the coarse matching set $C^{(1)}$. The geometric consistency score is defined by an initial transformation matrix $T_{init}=(R_{init},t_{init})$. For any given point pair $p_{\text{src},j}^{(1)}$ and $p_{\text{tar}, k}^{(1)}$, the consistency score is calculated as:
\begin{equation}
    S_{\text{geo}}^{(1)} \left( p_{\text{src}, j}^{(1)}, p_{\text{tar}, j}^{(1)} \right) 
    = \exp \Bigl( - \frac{\left\| R_{\text{init}} p_{\text{src}, j}^{(1)} + t_{\text{init}} - p_{\text{tar}, j}^{(1)} \right\|_2^2}{\sigma^2} \Bigr),
\end{equation}
where $\sigma$ is a scale parameter that modulates the sensitivity of the score. The use of the exponential function ensures that small misalignments are tolerated while large discrepancies lead to a rapid decline in the consistency score. 

Upon computing the geometric consistency scores for all point pairs $\bigl(p_{\text{src}, j}^{(1)}, p_{\text{tar}, j}^{(1)}\bigr) \in C^{(1)}$, the pairs are sorted in descending order based on their scores. Then retains the $k$ highest-scoring pairs, formalized as:
$\widetilde{C}^{(1)} = top$-$k\:\!\bigl(C^{(1)}, S_{\text{geo}}^{(1)}\bigr)$,
which refined subset $\widetilde{C}^{(1)}$ ensures that only the most geometrically consistent matches are preserved. This refined matching set serves as a robust foundation for subsequent fine registration processes by effectively filtering out outliers and erroneous matches. 

\subsection{Iterative Geometry-Aware Refinement Module}

The IGAR module functions within the fine matching phase, facilitating high-precision pose estimation through the integration of spatial geometric cues and an alternating optimization framework. \rev{Specifically, the module incorporates a Dynamic Geometric Consistency mechanism. This strategy prioritizes geometrically consistent pairs by adaptively weighting correspondences based on their spatial fidelity, thereby mitigating the influence of outliers. To quantify geometric reliability, let the source and target point clouds be denoted by the sets $P_{\mathrm{src}} = \{p_{\mathrm{src}, j}^{(1)}\}$ and $P_{\mathrm{tar}} = \{p_{\mathrm{tar}, j}^{(1)}\}$, initialized with a transformation $T_{\mathrm{init}}=(R_{\mathrm{init}},t_{\mathrm{init}})$.} We assign a consistency weight $w_{ij}$ to each correspondence pair, initially derived from the Euclidean spatial discrepancy:
\begin{equation}
w_{ij}^{(1)} = \exp \Bigl( - \frac{\| p_{\mathrm{src}, j}^{(1)} - p_{\mathrm{tar}, j}^{(1)} \|^2}{\sigma^2} \Bigr),
\end{equation}
where $\sigma^2$ denotes the scale parameter regulating sensitivity to spatial deviations. Through iterative refinement, both the weights and the transformation parameters are dynamically updated. \rev{At the $k$-th iteration, the weight is re-evaluated to reflect the instantaneous alignment error:
\begin{equation}
\begin{split}
w_{ij}^{(k)} ={} & \exp\Bigl(-\frac{\bigl\lVert p_{\mathrm{tar},j}^{(1)} - \bigl(R^{(k)} p_{\mathrm{src},j}^{(1)} + t^{(k)}\bigr)\bigr\rVert^2}{\sigma^2}\Bigr) \\
& \times \mathbb{I}\bigl[\bigl\lVert p_{\mathrm{tar},j}^{(1)} - \bigl(R^{(k)} p_{\mathrm{src},j}^{(1)} + t^{(k)}\bigr)\bigr\rVert < \tau \bigr],
\end{split}
\end{equation}
where $R^{(k)}$ and $t^{(k)}$ represent the optimized rotation and translation at iteration $k$, and $\mathbb{I}_{[\cdot]}$ serves as an indicator function that thresholds large deviations to exclude gross outliers. This formulation effectively embeds spatial distribution constraints into the optimization objective:
\begin{equation}
E(R, t) = \sum_{(i,j)\in \mathcal{C}} w_{ij}^{(k)} \| R p_{\mathrm{src},j}^{(1)} + t - p_{\mathrm{tar},j}^{(1)} \|_2^2.
\end{equation}}

To resolve this objective, we employ a weighted pseudo-center optimization strategy. The process commences by computing the weighted centroids of the source and target point sets, $\bar{p}_{\mathrm{src}}$ and $\bar{p}_{\mathrm{tar}}$, defined as:
\rev{\begin{equation}
\bar{p}_{\mathrm{src}}
= \frac{\displaystyle \sum_{(i,j)\in \mathcal{C}} w_{ij}^{(k)}\,p_{\mathrm{src},j}^{(1)}}
        {\displaystyle \sum_{(i,j)\in \mathcal{C}} w_{ij}^{(k)}}
\enspace,\enspace
\bar{p}_{\mathrm{tar}}
= \frac{\displaystyle \sum_{(i,j)\in \mathcal{C}} w_{ij}^{(k)}\,p_{\mathrm{tar},j}^{(1)}}
        {\displaystyle \sum_{(i,j)\in \mathcal{C}} w_{ij}^{(k)}}.
\end{equation}}

\rev{Subsequently, a weighted cross-covariance matrix $C \in \mathbb{R}^{3 \times 3}$ is constructed to encapsulate the structural correlation between the point sets, focusing the alignment on regions with high geometric consistency:
\begin{equation}
C
= \sum_{(i,j)\in \mathcal{C}} w_{ij}^{(k)}\,\bigl(p_{\mathrm{src},j}^{(1)} - \bar{p}_{\mathrm{src}}\bigr)\,\bigl(p_{\mathrm{tar},j}^{(1)} - \bar{p}_{\mathrm{tar}}\bigr)^{T}.
\end{equation}}

\rev{By performing SVD on the covariance matrix $C = U \Sigma V^T$, the optimal rotation matrix $R^*$ and translation vector $t^*$ are derived as:
\begin{equation}
R^* = VU^T, \quad t^* = \bar{p}_{\mathrm{tar}} - R^* \bar{p}_{\mathrm{src}}.
\end{equation}}

Ultimately, the IGAR module outputs a high-precision pose transformation matrix $T^*=[R^*, t^*]$. This transformation enables precise alignment of $P_{\mathrm{src}}$ with $P_{\mathrm{tar}}$, achieving robust and accurate registration.

\subsection{Loss function}
Our loss function $\mathcal{L}_{\text{total}}$ is composed of a matching loss function $\mathcal{L}_{\text{mat}}$, a keypoint loss function $\mathcal{L}_{\text{key}}$ and a dense loss function $\mathcal{L}_{\text{den}}$.

\textbf{Matching Loss.} Given a set $C$ of coarse correspondences, where each pair $(i,j)\in C$ is assigned an overlap weight $w_{ij}$, let $P^{(l)}_{ij}$ denote the matching probability. The hierarchical loss is formulated as:
\begin{equation}
\mathcal{L}_p = -\frac{1}{L} \sum_{l=1}^{L} \sum_{(i,j)\in C} w_{ij} \log \bigl(P_{ij}^{(l)}\bigr),
\end{equation}
which ensures that the predicted matching probabilities at multiple layers are consistent with the given overlap weights $w_{ij}$. This structure allows the network to learn representations at different scales and provides a gradual refinement of matches.
For the final matching probabilities $P_{ij}$, we enforce supervision using a weighted cross-entropy loss $\mathcal{L}_c$:

\begin{multline}
\mathcal{L}_{c} = - \frac{1}{\sum_{(i,j)\in C} \omega_{ij}} \sum_{(i,j)\in C} \omega_{ij} \log P_{ij} \\
\shoveleft{- \frac{1}{|N_X|} \sum_{k\in N_X} \log\Bigl(1 - \hat{\omega}^X_k\Bigr) + \frac{1}{|N_Y|} \sum_{k\in N_Y} \log\Bigl(1 - \hat{\omega}^Y_k\Bigr),}
\end{multline}
where $N_x$ and $N_y$ denote the sets of semi-dense nodes in the source and target clouds that are not involved in any correspondence, with $\hat{w}^X_k$ and $\hat{w}^Y_k$ representing their estimated overlap probabilities. The overlap weight $\omega_{ij}$ is computed based on a spherical neighborhood of radius $r$. The matching loss $\mathcal{L}_{\text{mat}}=\lambda_p \mathcal{L}_p + \lambda_c \mathcal{L}_c$, where $\lambda_p,\lambda_c$ are used to balance the terms. This dual aspect ensures that the network distinguishes between true matches and outliers, effectively reducing false correspondences.

\textbf{Keypoint Matching Loss.} For each valid keypoint correspondence $(x,x^+)$ with descriptors $d_x$ and $d_{x^+}$, we use a learnable weight matrix $W$ to compute similarity. An InfoNCE-style loss maximizes the similarity for true pairs while suppressing the negatives (denoted by the set $N_x$):

\begin{equation}
\begin{split}
\mathcal{L}_f &= -\mathbb{E}_{(x, x^+, N_x)}  \\
  &\Bigl[\log \frac{
  \exp\bigl(\langle d_x, W d_{x^+} \rangle\bigr)
  }{
  \exp\bigl(\langle d_x, W d_{x^+} \rangle\bigr)
  + \sum_{n \in N_x} \exp\bigl(\langle d_x, W d_n \rangle\bigr)
  } 
  \Bigr].
\end{split}
\end{equation}

For each keypoint $x$ with its predicted correspondence $\hat{y}$(extracted from its local patch $C_x$), 
we use an $\mathcal{L}_k$ loss to penalize positional errors:
\begin{equation}
\mathcal{L}_k = \mathbb{E}_{(x, \hat{y})}\bigl\|R\,x + t - \hat{y}\bigr\|^2.
\end{equation}

Each predicted correspondence is also assigned a confidence score $s$. We supervise these scores using a binary cross-entropy loss, where the ground truth label is 1 if $\| R x + t - \hat{y}\|\le\tau$. (with $\tau$ as a predefined threshold):
\begin{equation}
\mathcal{L}_i = -\mathbb{E}_{(x, \hat{y})} \left[ y \log s + (1 - y) \log (1 - s) \right].
\end{equation}

The keypoint matching loss function $\mathcal{L}_{\text{key}}= \lambda_f \mathcal{L}_f + \lambda_k \mathcal{L}_k + \lambda_i \mathcal{L}_i$.
The combination of these losses, each balanced by weights $\lambda_f,\lambda_k,\lambda_i$, ensures that both descriptor similarity and geometric accuracy are maintained.

\textbf{Dense Registration Loss.} To ensure the overall transformation is accurate, we supervise the dense registration phase by constraining the translation and rotation parameters:
$\mathcal{L}_t = \|\hat{t} - t\|^2,
\mathcal{L}_r = \|\hat{R}^T R - I\|_F^2,$
where $I$ is the $3*3$ identity matrix and the $F$ norm enforces closeness of the estimated rotation to the ground truth. The dense registration loss function $\mathcal{L}_{\text{den}}= \lambda_t \mathcal{L}_t + \lambda_r \mathcal{L}_r$, where $\lambda_t,\lambda_r$ are the weight of loss, guarantees that the final registration is not only locally accurate but also globally consistent.

The overall loss is a weighted combination of all the components: 
\begin{equation}
\mathcal{L}_{\text{total}} = \mathcal{L}_{\text{mat}}+\mathcal{L}_{\text{key}}+\mathcal{L}_{\text{den}}.
\end{equation}

This comprehensive loss formulation ensures robust training by addressing errors at various levels, including coarse matching, fine keypoint alignment, and dense global registration. By penalizing deviations at each stage, the proposed loss significantly enhances the accuracy and stability of the overall registration framework.

\begin{figure*}[t]
  \centering
  \includegraphics[width=1\linewidth]{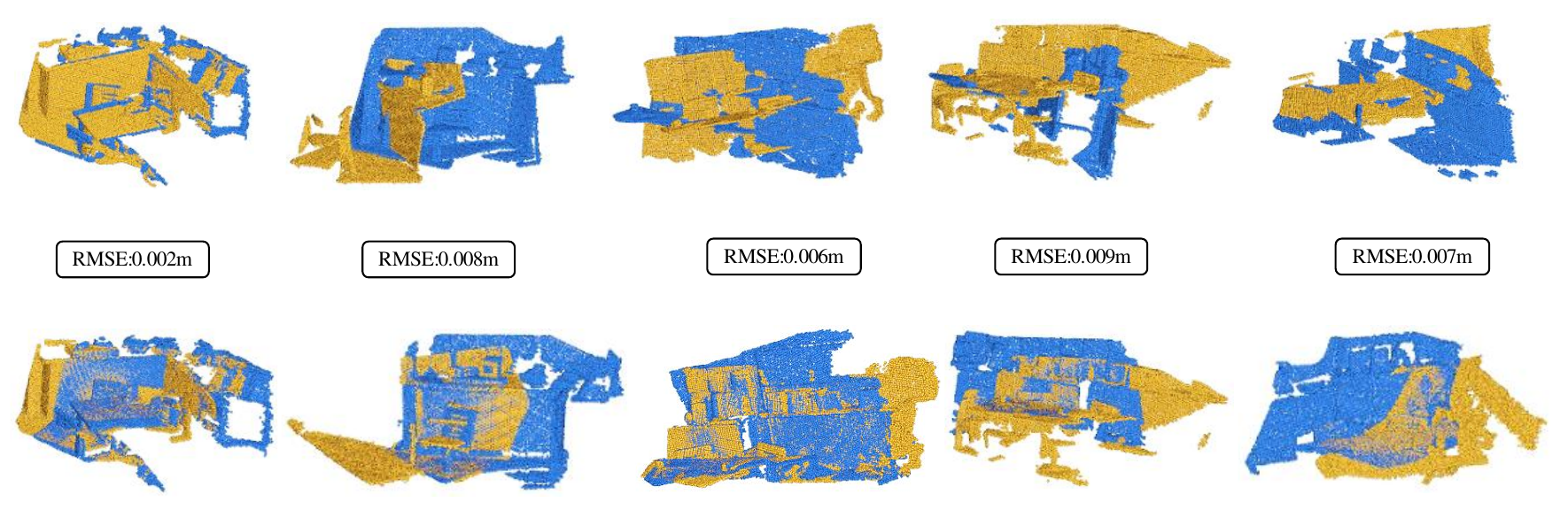}
  \caption{The result of IGASA on 3D(Lo)Match. We visualized some of the scenes in the dataset. The first row shows the initial arrangement of the two point clouds in their raw, unaligned positions.The second row displays the registration result predicted by our model.}
  \label{figurelabe4}
  \vspace{-5mm}
\end{figure*}

\section{EXPERIMENT}
\subsection{Experiment details}

\rev{The proposed IGASA framework is implemented in PyTorch. For the HPA module, the feature dimensions for the three pyramid levels ($F_{ordinary}, F_{minor}, F_{primary}$) are set to 64, 128, and 256, respectively. This progressive increase in channel depth ensures the capture of rich semantic abstractions. The kernel influence radius $R_{in}$ for KPConv is dynamically scaled as $2.5 \times dl$, where $dl$ is the voxel size at each level.
In the HCLA module, we employ a multi-head attention mechanism with 4 heads, where each head has a dimension of 64. The SGIRA and SAIGA units are stacked 1 time to balance computational efficiency with feature alignment performance.
For the IGAR module, we empirically set the number of iterations to $N=5$, which provides sufficient stability for the dynamic weight update mechanism without incurring excessive computational overhead.}

\rev{We evaluate our method on both indoor and outdoor benchmarks. For the 3DMatch and 3DLoMatch indoor datasets, raw point clouds are downsampled using a voxel size of 0.025m. To prevent overfitting, we apply data augmentation during training, including random rotations around arbitrary axes and random scaling. For the KITTI and nuScenes outdoor benchmarks, considering the sparsity of LiDAR data, we use a larger voxel size of 0.3m.}

\rev{The model is trained on a single NVIDIA RTX 3090 GPU using the AdamW optimizer with an initial learning rate of $10^{-4}$ and weight decay of $10^{-4}$. The learning rate is decayed by a factor of 0.95 every epoch. We train the model for 15 epochs on 3DMatch, 30 epochs on KITTI, and 10 epochs on nuScenes. The batch size is set to 1. For detailed loss weight configurations and specific augmentation parameters, please refer to the Supplementary Material}.


\begin{table*}[t]
\centering
\caption{Evaluation results of our method on 3DMatch and 3DLoMatch datasets. Optimal results are highlighted in bold, with runner-up results indicated by an underline.}
\label{table1}
\renewcommand{\arraystretch}{0.88}
\resizebox{0.8\textwidth}{!}{ 
\begin{tabular}{l|c|c|c|c|c|c|c|c|c|c}
\toprule 
\ &\multicolumn{5}{c|}{3DMatch} & \multicolumn{5}{c}{3DLoMatch} \\
\# Samples & \multicolumn{1}{c}{5000} & \multicolumn{1}{c}{2500} & \multicolumn{1}{c}{1000} & \multicolumn{1}{c}{500} & \multicolumn{1}{c|}{250} & \multicolumn{1}{c}{5000} & \multicolumn{1}{c}{2500} & \multicolumn{1}{c}{1000} & \multicolumn{1}{c}{500} & 250 \\
\midrule
\multicolumn{11}{c}{\textit{Feature Matching Recall (\%) }$\uparrow$} \\
\midrule
FCGF~\cite{choy2019fully} & 97.4 & 97.3 & 97.0 & 96.7 & 96.6 & 76.6 & 75.4 & 74.2 & 71.7 & 67.3 \\
Predator~\cite{huang2021predator} & 96.6 & 96.6 & 96.5 & 96.3 & 95.3 & 78.6 & 76.7 & 75.7 & 75.7 & 72.9 \\
YOHO~\cite{wang2022you} & \textbf{98.2} & 97.6 & 97.5 & 97.7 & 96.0 & 79.4 & 78.1 & 76.3 & 73.8 & 69.1 \\
CoFiNet~\cite{yu2021cofinet} & \underline{98.1} & \underline{98.3} & \underline{98.1} & 98.2 & 98.3 & 83.1 & 83.5 & 83.3 & 83.1 & 82.6 \\
GeoTransformer~\cite{qin2022geometric} & 97.9 & 97.9 & 97.9 & 97.9 & 97.6 & 88.3 & 88.6 & 88.8 & \underline{88.6} & \underline{88.3} \\
PCR-CG~\cite{zhang2022pcr} & 97.4 & 97.5 & 97.7 & 97.3 & 97.6 & 80.4 & 82.2 & 82.6 & 83.2 & 82.8 \\
OIF-Net~\cite{yang2022one} & \underline{98.1} & 98.1 & 97.9 & \underline{98.4} & \underline{98.4} & 84.6 & 85.2 & 85.5 & \underline{88.6} & 87.0 \\
RoITr~\cite{yu2023rotation} & 98.0 & 98.0 & 97.9 & 98.0 & 97.9 & \textbf{89.6} & \textbf{89.6} & \textbf{89.5} & \textbf{89.4} & \textbf{89.3} \\
SIRA-PCR~\cite{chen2023sira} & \textbf{98.2} & \textbf{98.4} & \textbf{98.4} & \textbf{98.5} & \textbf{98.5} & \underline{88.8} & \underline{89.0} & \underline{88.9} & \underline{88.6} & 87.7 \\
IGASA (\textit{ours}) & \textbf{98.2} & 97.8 & 97.8 & 97.9 & 97.8 & 82.1 & 80.5 & 80.5 & 80.5 & 80.4 \\
\midrule
\multicolumn{11}{c}{\textit{Registration Recall (\%)} $\uparrow$} \\
\midrule
FCGF~\cite{choy2019fully} & 85.1 & 83.4 & 82.7 & 81.6 & 76.6 & 74.4 & 70.4 & 67.2 & 63.4 & 61.4 \\
Predator~\cite{huang2021predator} & 89.0 & 89.9 & 90.6 & 88.5 & 86.6 & 61.2 & 61.2 & 62.4 & 60.8 & 58.1 \\
YOHO~\cite{wang2022you} & 90.8 & 90.3 & 89.1 & 88.6 & 84.5 & 67.5 & 68.0 & 62.2 & 63.4 & 61.9 \\
CoFiNet~\cite{yu2021cofinet} & 92.0 & 91.8 & 91.8 & 91.4 & 91.2 & 75.0 & \underline{74.8} & 74.2 & 74.1 & 73.5 \\
GeoTransformer~\cite{qin2022geometric} & 92.0 & 91.8 & 91.1 & 91.4 & 92.0 & \underline{75.5} & \underline{74.8} & 74.6 & \underline{74.6} & \underline{75.3} \\
PCR-CG~\cite{zhang2022pcr} & 89.4 & 90.7 & 90.0 & 88.7 & 86.8 & 66.3 & 67.2 & 69.0 & 68.5 & 65.0 \\
OIF-Net~\cite{yang2022one} & 92.4 & 91.9 & 91.8 & 91.4 & 91.0 & 74.7 & \underline{74.8} & \underline{74.8} & 74.2 & 73.6 \\
RoITr~\cite{yu2023rotation} & 91.9 & 91.7 & 91.8 & 91.4 & 91.0 & 74.7 & \underline{74.8} & \underline{74.8} & 74.2 & 73.6 \\
\rev{FeatSync~\cite{hu2024featsync}} & 90.2 & 90.3 & 90.6 & 91.2 & 91.1 & - & - & - & - & - \\
SIRA-PCR~\cite{chen2023sira} & \underline{93.6} & \underline{93.9} & \underline{93.9} & \underline{92.7} & \underline{92.4} & 73.5 & 73.9 & 73.0 & 73.4 & 71.1 \\
IGASA (\textit{ours}) & \textbf{94.6} & \textbf{94.5} & \textbf{94.5} & \textbf{94.4} & \textbf{94.3} & \textbf{76.5} & \textbf{76.5} & \textbf{76.3} & \textbf{76.4} & 
\textbf{76.4} \\
\midrule
\multicolumn{11}{c}{\textit{Inlier Ratio (\%) }$\uparrow$} \\
\midrule
FCGF~\cite{choy2019fully} & 56.8 & 54.1 & 48.7 & 42.5 & 34.1 & 21.4 & 20.0 & 17.2 & 14.8 & 11.6 \\
Predator~\cite{huang2021predator} & 58.0 & 58.4 & 57.1 & 54.1 & 49.3 & 26.7 & 28.1 & 28.3 & 27.5 & 25.8 \\
YOHO~\cite{wang2022you} & 64.4 & 60.7 & 55.7 & 46.4 & 41.2 & 25.9 & 23.3 & 22.6 & 18.2 & 15.0 \\
CoFiNet~\cite{yu2021cofinet} & 49.8 & 51.2 & 51.9 & 52.2 & 52.2 & 24.4 & 25.9 & 26.7 & 26.8 & 26.9 \\
GeoTransformer~\cite{qin2022geometric} & 71.9 & 75.2 & 76.0 & 82.2 & 85.1 & 43.5 & 45.3 & 46.2 & 52.9 & 57.7 \\
OIF-Net~\cite{yang2022one} & 62.3 & 65.2 & 66.8 & 67.1 & 67.5 & 27.5 & 30.0 & 31.2 & 32.6 & 33.1 \\
RoITr~\cite{yu2023rotation} & \underline{82.6} & \underline{82.8} & 83.0 & 83.0 & 83.0 & \underline{54.3} & \underline{54.6} & 55.1 & 55.2 & 55.3 \\
\rev{OT-CA~\cite{guo2024learning}} & 59.5 & 61.5 & 60.9 & 61.7 & 61.7 & 30.5 & 30.6 & 30.5 & 34.9 & 30.2 \\
SIRA-PCR~\cite{chen2023sira} & 70.8 & 78.3 & \underline{83.7} & \underline{85.9} & \underline{87.4} & 43.3 & 49.0 & \underline{55.9} & \underline{58.8} & \underline{60.7} \\
IGASA (\textit{ours}) & \textbf{87.9} & \textbf{87.8} & \textbf{87.9} & \textbf{87.9} & \textbf{87.9} & \textbf{61.6} & \textbf{61.5} & \textbf{61.5} & \textbf{61.6} & \textbf{61.4} \\
\bottomrule 
\end{tabular}
}
\vspace{-5mm}
\end{table*}

\subsection{Indoor Benchmarks: 3DMatch \& 3DLoMatch}

\textbf{Dataset.} The 3DMatch~\cite{zeng20173dmatch} and 3DLoMatch~\cite{huang2021predator} datasets are constructed from RGB-D data. The 3DMatch dataset contains 62 indoor scenes, with 46 scenes allocated for training, 8 for validation, and 8 for testing. Each scene in the 3DMatch dataset consists of point cloud pairs with at least a 30\% overlap. In contrast, the point clouds in the 3DLoMatch dataset have a smaller overlap, ranging from 10\% to 30\%.

\textbf{Metrics.} In our evaluation, we employ three key metrics to comprehensively assess registration performance: Inlier Ratio (IR), Feature Matching Recall (FMR), and Registration Recall (RR). The IR quantifies the proportion of candidate correspondences with residual errors below a set threshold ($e.g.,$ 0.1 m) under the ground-truth transformation. FMR measures the fraction of point cloud pairs that achieve an inlier ratio above a designated threshold ($e.g.,$ 5\%).

\textbf{Registration results.} In summary, IGASA outperforms state-of-the-art methods on both datasets, excelling in RR, IR, and FMR. \rev{We compare our method IGASA, with recent methods: 
FCGF~\cite{choy2019fully},  Predator~\cite{huang2021predator}, YOHO~\cite{wang2022you}, CoFiNet~\cite{yu2021cofinet}, GeoTransformer~\cite{qin2022geometric}, PCR-CG~\cite{zhang2022pcr}, OIF-Net~\cite{yang2022one}, RoITr~\cite{yu2023rotation}, FeatSync~\cite{hu2024featsync}, OT-CA~\cite{guo2024learning} and SIRA-PCR~\cite{chen2023sira}. The outcomes for 3DMatch and 3DLoMatch are displayed in \cref{table1}. Our method demonstrates significant improvements in multiple evaluation metrics. For a more comprehensive evaluation, please refer to Table S.IV in the Supplementary Material.}

To 3DMatch dataset, the results indicate that IGASA achieves a competitive FMR; for instance, at 5000 samples, it attains an FMR of 98.2\%, comparable to YOHO~\cite{wang2022you} and SIRA‐PCR~\cite{chen2023sira}. However, under sparser conditions ($e.g.,$ 250 or 500 samples), IGASA’s FMR is slightly lower than some competing methods, suggesting that feature robustness under limited samples may be an area for further improvement.
In terms of RR, IGASA consistently outperforms the compared methods, recording values of 94.6\%, 94.5\%, 94.5\%, 94.4\%, and 94.3\% as the sample size decreases from 5000 to 250.
Additionally, IGASA demonstrates stable performance in the IR metric, maintaining 87.9\% across most sample sizes. This result is notably higher than those of methods such as GeoTransformer~\cite{qin2022geometric}, RoITr~\cite{yu2023rotation}, and SIRA‐PCR~\cite{chen2023sira}, reflecting robust inlier extraction and enhanced noise resistance.

The results in \cref{table1} show that IGASA performs well on the 3DLoMatch dataset across key metrics: FMR, RR and IR. In RR, IGASA performs best among all methods, with a peak of 76.5\% at 5000 and 2500 samples, surpassing competitors like RoITr~\cite{yu2023rotation} and SIRA-PCR~\cite{chen2023sira}. Additionally, in terms of IR, IGASA leads with 61.6\% at 5000 samples and maintains high values across smaller sample sizes, demonstrating superior inlier extraction.
Overall, these results demonstrate IGASA's potential to outperform traditional point cloud registration methods.

\textbf{Qualitative results.} \cref{figurelabe4} provides a series of qualitative registration results on 3DM(Lo)atch, which emphasize the advantages of IGASA in terms of registration results.

\subsection{Outdoor Benchmark: KITTI odometry}

\textbf{Dataset.} KITTI~\cite{geiger2013vision} is a widely recognized benchmark for autonomous driving, consisting of 11 outdoor driving sequences captured by LiDAR sensors. Following common protocols established in prior work, sequences 0-5 are used for training, 6-7 for validation, and 8-10 for testing. 

\begin{table}[t]
\centering
\caption{Registration performance on KITTI dataset.}
\label{table3}
\resizebox{0.8\columnwidth}{!}{ 
\begin{tabular}{lccc}
\toprule
Model & RTE (cm) & RRE (°) & RR (\%) \\
\midrule
3DFeat-Net~\cite{yew20183dfeat}    & 25.9 & 0.25 & 96.0  \\
FCGF~\cite{choy2019fully}           & 9.5  & 0.30 & 96.6  \\
Predator~\cite{huang2021predator}       & 6.8  & 0.27 & \underline{99.8} \\
CoFiNet~\cite{yu2021cofinet}        & 8.2  & 0.41 & \underline{99.8} \\
HRegNet~\cite{lu2021hregnet}       & 12  & 0.29 & 99.7 \\
GeoTransformer~\cite{qin2022geometric}  & 6.8  & \underline{0.24} & \underline{99.8} \\
OIF-Net~\cite{yang2022one}         & \underline{6.5}  & \textbf{0.23} & \underline{99.8} \\
MAC~\cite{zhang20233d}            & 8.5  & 0.40 & 99.5 \\
IGASA (\textit{ours})           & \textbf{4.6}  & \underline{0.24} & \textbf{100.0} \\
\bottomrule
\end{tabular}
}
\vspace{-4mm}
\end{table}

\begin{figure*}[t]
  \centering
  \includegraphics[width=1.0\linewidth]{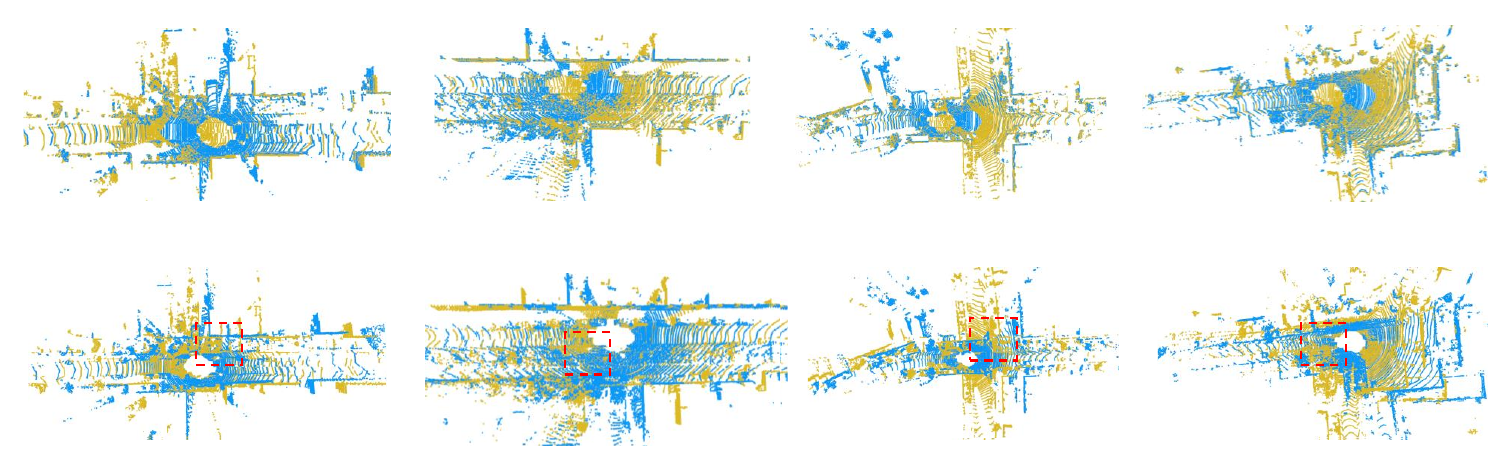}
  \caption{\rev{Qualitative registration results on the KITTI odometry benchmark. We visualize four representative driving scenes characterized by large spatial extents and varying structural distinctiveness. The first row displays the initial unaligned input pairs (Source: Blue, Target: Yellow), illustrating significant initial pose discrepancies. The second row shows the registration results achieved by IGASA. The red rectangular boxes highlight specific regions with sharp structural boundaries, such as road curbs and intersection corners. Note that despite the limited overlap and noise typical of outdoor LiDAR scans, our method precisely aligns fine-grained structures such as trees, vehicles, and road boundaries.}}
  \label{figurelabe5}
  \vspace{-5mm}
\end{figure*}

\textbf{Metrics.} To evaluate the performance, we adopt three crucial metrics. RR, Relative Rotation Error (RRE), and Relative Translation Error (RTE). RR quantifies the proportion of point cloud pairs for which the estimated transformation meets a predefined accuracy threshold. Meanwhile, RRE measures the angular discrepancy between the estimated and ground-truth rotations, and RTE evaluates the distance between the estimated and true translation vectors.

\textbf{Registration Results.} For the KITTI dataset evaluation, as shown in Table \ref{table3}, outperforms all state-of-the-art methods, \rev{including
3DFeat-Net~\cite{yew20183dfeat}, FCGF~\cite{choy2019fully}, Predator~\cite{huang2021predator}, CoFiNet~\cite{yu2021cofinet}, HRegNet~\cite{lu2021hregnet}, GeoTransformer~\cite{qin2022geometric}, OIF-Net~\cite{yang2022one} and MAC~\cite{zhang20233d}.} Specifically, IGASA achieves an RTE of 4.6 cm, an RRE of 0.24°, and an RR of 100.0\%, which are all the best results among total compared methods. 

Compared to traditional feature-based methods like 3DFeat-Net~\cite{yew20183dfeat} and FCGF~\cite{choy2019fully}, IGASA shows significant improvements. While 3DFeat-Net~\cite{yew20183dfeat} has an RTE of 25.9 cm and a RR of 96.0\%, IGASA's performance in both accuracy and registration recall is superior, demonstrating the strength of the model in handling complex registration tasks.
IGASA also leads the comparison with transformer-based models like GeoTransformer~\cite{qin2022geometric} and OIF-Net~\cite{yang2022one}, achieving higher RR and lower RTE.

\textbf{Qualitative results.} \rev{To further validate the robustness of IGASA in outdoor environments, we present qualitative visualization results on the KITTI dataset in \cref{figurelabe5}. As shown in the second row, our method achieves precise alignment even in scenes with large initial misalignment and repetitive geometric structures. Highlighted by the red rectangular boxes, the tight overlap between the source (blue) and target (yellow) point clouds demonstrates the high accuracy of our estimated 6-DoF pose, particularly around road curbs and structural corners. This visual alignment confirms the effectiveness of the proposed method in capturing fine geometric details.}

\begin{table}[t]
\centering
\caption{Registration performance on nuScenes dataset.}
\label{table4}
\resizebox{0.85\columnwidth}{!}{ 
\begin{tabular}{lccc}
\toprule
Model & RTE (m) & RRE (°) & RR (\%) \\
\midrule
FGR~\cite{zhou2016fast}         & 0.71  & 1.01 & 32.2  \\
DCP~\cite{wang2019deep}       & 1.09  & 2.07 & 56.8 \\
IDAM~\cite{li2020iterative}        & 0.47  & 0.79 & 88.0 \\
FMR~\cite{huang2020feature}       & 0.60  & 1.61 & 92.1 \\
DGR~\cite{choy2020deep}  & 0.21  & 0.48 & \underline{98.4} \\
HRegNet~\cite{lu2021hregnet}      & \underline{0.18}  & \underline{0.45} & \textbf{99.9} \\
IGASA (\textit{ours})           & \textbf{0.12}  & \textbf{0.21} & \textbf{99.9} \\
\bottomrule
\end{tabular}
}
\vspace{-4mm}
\end{table}

\subsection{Outdoor Benchmark: nuScenes odometry}
\textbf{Dataset.} NuScenes~\cite{caesar2020nuscenes} is a large-scale dataset designed for autonomous driving, capturing diverse outdoor scenarios with data from a complete suite of sensors. It comprises 850 scenes for training and validation, with an additional 150 scenes reserved for testing. We partition the dataset by assigning the first 700 scenes for training while using the remainder for validation. To optimize the processing of raw data, we apply a voxel grid down-sampling with a resolution of 0.3 m, which effectively reduces point cloud density while preserving essential geometric details. 

\textbf{Metrics.} We adopt RR, RRE and RTE to thoroughly assess registration performance.

\textbf{Registration Results.} For LiDAR benchmark nuScenes, we evaluate our IGASA by comparing with several recent state-of-the-art methods~\cite{lu2021hregnet},~\cite{zhou2016fast},~\cite{wang2019deep},~\cite{li2020iterative,huang2020feature},~\cite{choy2020deep} as shown in Table \ref{table4}. The experimental results demonstrate that IGASA outperforms competing methods in point cloud registration. IGASA achieves an RTE of 0.12 m, an RRE of 0.21° and a RR of 99.9\%, surpassing traditional methods. 

\begin{figure*}[t]
  \centering
  \includegraphics[width=1.0\linewidth]{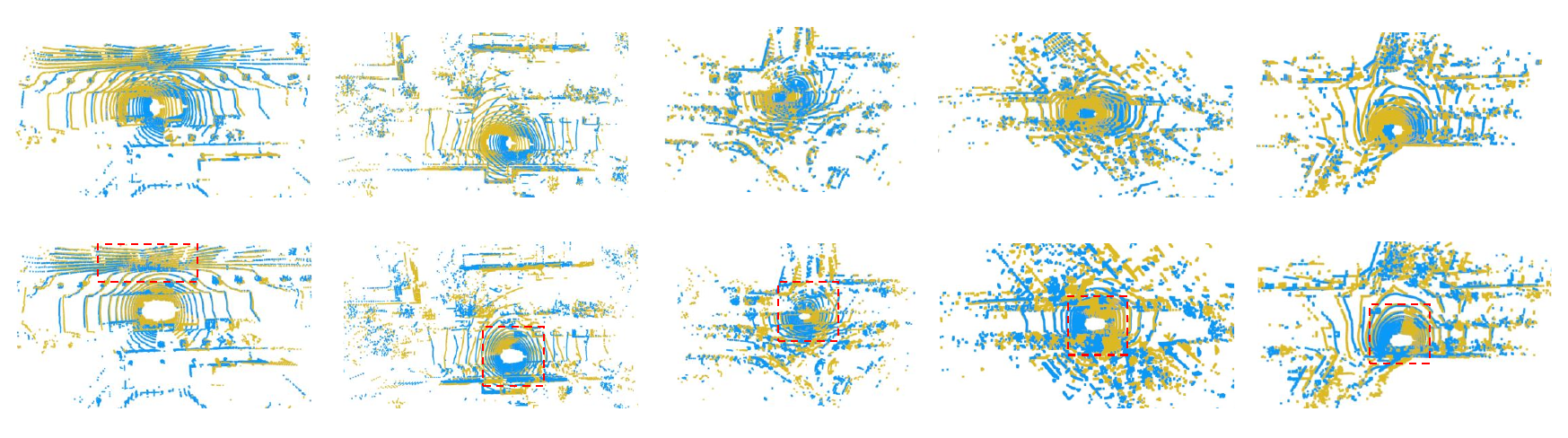}
  \caption{\rev{Qualitative registration results on the nuScenes dataset. These scenes exhibit extreme sparsity and non-uniform point density, which pose significant challenges for correspondence estimation. The first row shows the raw, unaligned point clouds. The second row presents the aligned results from IGASA. The red rectangular boxes highlight specific regions containing aligned sparse geometric cues, such as ground scan lines and structural edges. It can be observed that IGASA robustly handles the sparse geometry, effectively aligning the ground plane and sparse vertical structures ($e.g.,$ traffic signs), demonstrating its efficacy in low-texture environments.}}
  \label{figurelabe6}
  \vspace{-5mm}
\end{figure*}

\textbf{Qualitative results.} \rev{\cref{figurelabe6} illustrates the registration performance on the nuScenes dataset, which is characterized by sparse LiDAR points and low overlap. Despite the lack of dense geometric features, IGASA successfully recovers the correct transformation parameters. As highlighted by the red rectangular boxes in the bottom row, the aligned results demonstrate that our HCLA module effectively captures global context from sparse data, ensuring reliable registration even when local features are ambiguous or missing.}

\subsection{Ablation Experiment}

\subsubsection{Related Modules in IGASA}

\rev{To assess the individual contribution of each component, we conducted a comprehensive ablation study on the 3DMatch dataset, as summarized in \cref{table5}. The baseline model, which utilizes a standard UNet-like backbone without the proposed attention and refinement modules, yields a RR of only 89.6\%. The integration of the HCLA module significantly boosts the recall to 92.8\%, a substantial improvement that validates the importance of bridging the semantic gap. This gain indicates that the SGIRA and SAIGA units effectively guide the feature learning process, ensuring that local geometry is semantically aligned before matching.}

\rev{Most notably, the addition of the IGAR module further elevates the performance to 94.6\%. This improvement is particularly evident in the IR, which rises from 79.2\% to 87.9\%. This finding underscores the critical role of the iterative refinement stage. While HCLA provides a robust initial alignment, it is the dynamic geometric consistency weighting within IGAR that effectively suppresses residual outliers and fine-tunes the pose, ensuring high-precision registration even in complex scenes. The complete IGASA framework thus achieves the optimal balance between robust feature extraction and precise geometric alignment.}





\begin{table}[t]
\centering
\caption{\rev{Ablation study of key modules in the IGASA framework on the 3DMatch and 3DLoMatch datasets. The results are reported in percentage (\%). Best performance is highlighted in bold. The baseline denotes the network without HCLA and IGAR.}
}
\resizebox{0.9\columnwidth}{!}{ 
\label{table5}
\begin{tabular}{ccc|ccc|ccc}
\toprule
\multicolumn{3}{c}{Module} & \multicolumn{3}{c}{3DMatch} & \multicolumn{3}{c}{3DLoMatch} \\
HPA & HCLA & IGAR & FMR & IR & RR & FMR & IR & RR \\
\midrule
$\checkmark$ & - & - & 92.7 & 80.2  & 91.3 & 72.8 & 49.7 & 70.8 \\ 
- & $\checkmark$ & - & 94.2 & 83.5  & 91.9 & 75.9 & 55.1 & 72.7 \\ 
- & - & $\checkmark$ & 93.0 & 81.7  & 91.5 & 69.2 & 52.6  & 71.5\\
- & $\checkmark$ & $\checkmark$ & 97.2 & 85.2 & 93.4 & 74.3 & 57.6 & 74.2  \\
$\checkmark$ & - & $\checkmark$ & 95.6 & 81.9 & 92.8 & 71.0 & 51.3 & 72.3  \\
$\checkmark$ & $\checkmark$ & - & 96.9 & 83.7 & 93.2 & 78.9 & 53.8 & 73.5 \\
$\checkmark$ & $\checkmark$ & $\checkmark$  & \textbf{98.2} & \textbf{87.9} & \textbf{94.6} & \textbf{82.1} & \textbf{61.6} & \textbf{76.5}  \\
\bottomrule
\end{tabular}
}
\vspace{-5mm}
\end{table}

\subsubsection{SGIRA and SAIGA in HCLA Module}

\rev{We further investigate the internal mechanisms of the HCLA module by analyzing the separate contributions of its two sub-components: SGIRA and SAIGA. The quantitative results are presented in Table \ref{table6}.}

\rev{The results reveal a clear hierarchy in feature enhancement. Activating only the SGIRA module improves the FMR from 95.6\ to 96.2\%. This gain is attributed to the semantic filtering capability of SGIRA, which leverages global context to suppress ambiguous local features. Subsequently, enabling only SAIGA yields a further improvement in IR (84.2\%), suggesting that the intrinsic geometric self-attention successfully reinforces spatial distinctiveness, making the descriptors more robust to viewpoint changes.}

\rev{The peak performance is achieved when both components are active (Row 4), yielding an FMR of 98.2\% and an IR of 87.9\%. This significant boost confirms the synergistic relationship between the two units: SGIRA establishes a semantic consensus to filter noise, while SAIGA refines the local geometric details. Together, they ensure that the generated correspondences are both semantically consistent and geometrically precise.}




\begin{table}[t]
\centering
\scriptsize 
\caption{Ablation experiments of the tow components in HCLA module on 3DMatch dataset. The results are reported in percentage (\%). Best performance is highlighted in bold.}
\resizebox{0.65\columnwidth}{!}{ 
\label{table6}
\begin{tabular}{cc|c|c|c}
\toprule
\multicolumn{2}{c}{Module} & \multicolumn{3}{c}{3DMatch}  \\
SGIRA & SAIGA & FMR & IR & RR  \\
\midrule
- & - & 95.6 & 81.9 & 92.8  \\ 
$\checkmark$ & - & 96.2 & 83.3 & 93.2  \\ 
- & $\checkmark$ & 96.7 & 84.2 & 93.5  \\ 
$\checkmark$ & $\checkmark$ & \textbf{98.2} & \textbf{87.9} & \textbf{94.6}  \\
\bottomrule
\end{tabular}
}
\vspace{-5mm}
\end{table}

\subsubsection{Analysis of Loss Functions}
\rev{The optimization of IGASA relies on a composite objective function. To validate the necessity of each constraint, we evaluate the performance using combinations of the matching loss ($\mathcal{L}_{\text{mat}}$), keypoint loss ($\mathcal{L}_{\text{key}}$), and dense registration loss ($\mathcal{L}_{\text{den}}$). The results are summarized in Table \ref{table7}.}

\rev{Using any single loss function results in suboptimal performance, indicating that individual constraints are insufficient for the full registration pipeline. For instance, relying solely on $\mathcal{L}_{\text{den}}$ yields a lower IR (71.5\%), as it lacks explicit supervision for coarse correspondence establishment. The combination of $\mathcal{L}_{\text{mat}}$ and $\mathcal{L}_{\text{den}}$ provides a significant boost, highlighting the importance of simultaneously supervising the coarse matching probability and the fine-grained spatial distance.The best results are obtained when all three loss terms are jointly optimized. The inclusion of $\mathcal{L}_{\text{key}}$ further refines the metric by guiding the network to focus on salient overlapping regions. This multi-task supervision strategy ensures that the network learns robust features (via $\mathcal{L}_{\text{mat}}$), identifies reliable keypoints (via $\mathcal{L}_{\text{key}}$), and achieves precise spatial alignment (via $\mathcal{L}_{\text{den}}$), thereby maximizing registration accuracy.}




\begin{table}[t]
\centering
\caption{\rev{Impact of different loss function combinations on registration performance. $\mathcal{L}_{\text{mat}}$, $\mathcal{L}_{\text{key}}$, and $\mathcal{L}_{\text{den}}$ denote the matching, keypoint, and dense registration losses, respectively. The results are reported in percentage (\%). Best performance is highlighted in bold.}}
\label{table7}
\begin{tabular}{cccccc}
\toprule
$\mathcal{L}_{\text{mat}}$&$\mathcal{L}_{\text{key}}$&$\mathcal{L}_{\text{den}}$ & FMR & IR & RR  \\ 
\midrule
$\checkmark$ & - & - & 93.4 & 74.9 & 91.5  \\ 
- & $\checkmark$ & - & 92.8 & 72.2 & 90.3  \\ 
- & - & $\checkmark$ & 93.0 & 71.5 & 90.7  \\
- & $\checkmark$ & $\checkmark$ & 95.8 & 72.5 & 91.3  \\
$\checkmark$ & - & $\checkmark$ & 96.4 & 75.6 & 92.4  \\
$\checkmark$ & $\checkmark$ & - & 96.3 & 77.1 & 92.1  \\
$\checkmark$ & $\checkmark$ & $\checkmark$  & \textbf{98.2} & \textbf{87.9} & \textbf{94.6}  \\
\bottomrule
\end{tabular}
\vspace{-5mm}
\end{table}

\begin{table}[t]
\centering
\caption{\rev{Computational efficiency comparison on the 3DMatch benchmark. Times are measured in seconds (s). The total time represents the sum of feature extraction (Model time) and transformation estimation (Pose time).}}
\label{table8}
\begin{tabular}{lccc}
\toprule
Methods & Model time & Pose time & Total time \\
\midrule
FCGF~\cite{choy2019fully}            & 0.073           & 4.332          & 4.405          \\
D3Feat~\cite{bai2020d3feat}          & 0.035           & 4.472          & 4.506          \\
SpinNet~\cite{ao2021spinnet}         & 84.569          & 3.583          & 88.152         \\
Predator~\cite{huang2021predator}        & \textbf{0.050}  & 7.014          & 7.064          \\
CoFiNet~\cite{yu2021cofinet}         & 0.167           & \textbf{2.493} & \textbf{2.660} \\
GeoTransformer~\cite{qin2022geometric}  & 0.124           & 2.557          & 2.701          \\
HECPG~\cite{xie2024hecpg}        & 0.112           & 2.584          & 2.696          \\
IGASA(\textit{ours}) &0.135&2.628&2.763 \\
\bottomrule
\end{tabular}
\vspace{-5mm}
\end{table}

\subsubsection{\rev{Efficiency Analysis}}


\rev{To evaluate the practical applicability of the proposed framework, we conducted a runtime analysis on the 3DMatch benchmark, comparing IGASA against key baselines and recent state-of-the-art methods~\cite{huang2021predator,ao2021spinnet,qin2022geometric,yu2021cofinet}. The results, summarized in Table \ref{table8}, decompose the total inference time into feature extraction (Model time) and transformation estimation (Pose time).}

\rev{The comparison reveals that classical methods often suffer from extreme computational bottlenecks; for instance, SpinNet requires over 88 seconds per frame due to its voxelization process, and Predator, while fast in feature extraction, incurs a high pose estimation cost (7.014 s) due to its reliance on heavy RANSAC iterations. In contrast, IGASA demonstrates high efficiency with a total inference time of 2.763 s. This performance is highly competitive with leading transformer-based methods such as GeoTransformer (2.701 s) and CoFiNet (2.660 s), and is comparable to the HECPG (2.696 s).}

\rev{Although IGASA incurs a marginal computational overhead of approximately 0.1 s compared to CoFiNet, this increase is attributable to the additional computations required by the dual-unit HCLA module and the iterative refinement in IGAR. Given the significant gains in registration recall and inlier ratio demonstrated in previous sections, this slight increase in latency is a justifiable trade-off. The results confirm that IGASA effectively balances model complexity and execution speed, making it suitable for applications requiring both high precision and reasonable time efficiency.}

\section{Conclusion}


\rev{In this paper, we presented IGASA as an integrated framework for robust point cloud registration through the synergy of geometry-aware refinement and skip-attention mechanisms. The proposed Hierarchical Pyramid Architecture (HPA) provides a structured foundation for multi-scale feature extraction. Within this architecture, the Hierarchical Cross-Layer Attention (HCLA) module effectively bridges the semantic gap by aligning multi-resolution features and enhancing both global semantics and local geometric consistency. Furthermore, the Iterative Geometry-Aware Refinement (IGAR) module improves registration precision by leveraging spatial geometric cues to dynamically suppress outliers through an alternating optimization strategy.}
Extensive evaluations conducted on the 3D(Lo)Match, KITTI, and nuScenes datasets demonstrate that our method consistently outperforms state-of-the-art techniques in terms of accuracy and robustness. These results highlight the efficacy of combining multi-resolution feature fusion with iterative geometric refinement for complex 3D vision tasks. \rev{However, the framework exhibits a slight increase in computational latency due to the iterative nature of the refinement process.}
In the future, we aim to further optimize the framework to enhance its adaptability to highly dynamic environments~\cite{wang2023enhancing,li2023synergy} and explore more efficient architectures for real-time large-scale point cloud processing.


\bibliographystyle{ieeetr}
\bibliography{main}

@String(ICCV= {Int. Conf. Comput. Vis.})

@String(ECCV= {Eur. Conf. Comput. Vis.})

@String(ICCV  = {ICCV})

@String(ECCV  = {ECCV})

@inproceedings{yew20183dfeat,
  title={3dfeat-net: Weakly supervised local 3d features for point cloud registration},
  author={Yew, Zi Jian and Lee, Gim Hee},
  booktitle={Proceedings of the European conference on computer vision (ECCV)},
  pages={607--623},
  year={2018}
}

@inproceedings{zeng20173dmatch,
  title={3dmatch: Learning local geometric descriptors from rgb-d reconstructions},
  author={Zeng, Andy and Song, Shuran and Nie{\ss}ner, Matthias and Fisher, Matthew and Xiao, Jianxiong and Funkhouser, Thomas},
  booktitle={Proceedings of the IEEE conference on computer vision and pattern recognition},
  pages={1802--1811},
  year={2017}
}

@inproceedings{huang2024kdd,
  title={Kdd-loam: Jointly learned keypoint detector and descriptors assisted lidar odometry and mapping},
  author={Huang, Renlang and Zhao, Minglei and Chen, Jiming and Li, Liang},
  booktitle={2024 IEEE International Conference on Robotics and Automation (ICRA)},
  pages={8559--8565},
  year={2024},
  organization={IEEE}
}

@inproceedings{wang2021deep,
  title={Deep two-view structure-from-motion revisited},
  author={Wang, Jianyuan and Zhong, Yiran and Dai, Yuchao and Birchfield, Stan and Zhang, Kaihao and Smolyanskiy, Nikolai and Li, Hongdong},
  booktitle={Proceedings of the IEEE/CVF conference on Computer Vision and Pattern Recognition},
  pages={8953--8962},
  year={2021}
}

@article{zhang2024comprehensive,
  title={A comprehensive survey and taxonomy on point cloud registration based on deep learning},
  author={Zhang, Yu-Xin and Gui, Jie and Cong, Xiaofeng and Gong, Xin and Tao, Wenbing},
  journal={arXiv preprint arXiv:2404.13830},
  year={2024}
}

@inproceedings{wen2020point,
  title={Point cloud completion by skip-attention network with hierarchical folding},
  author={Wen, Xin and Li, Tianyang and Han, Zhizhong and Liu, Yu-Shen},
  booktitle={Proceedings of the IEEE/CVF conference on computer vision and pattern recognition},
  pages={1939--1948},
  year={2020}
}

@article{chen2019multi,
  title={Multi-modal fusion network with multi-scale multi-path and cross-modal interactions for RGB-D salient object detection},
  author={Chen, Hao and Li, Youfu and Su, Dan},
  journal={Pattern Recognition},
  volume={86},
  pages={376--385},
  year={2019},
  publisher={Elsevier}
}

@inproceedings{agarwal2023attention,
  title={Attention attention everywhere: Monocular depth prediction with skip attention},
  author={Agarwal, Ashutosh and Arora, Chetan},
  booktitle={Proceedings of the IEEE/CVF Winter Conference on Applications of Computer Vision},
  pages={5861--5870},
  year={2023}
}

@article{yu2021cofinet,
  title={Cofinet: Reliable coarse-to-fine correspondences for robust pointcloud registration},
  author={Yu, Hao and Li, Fu and Saleh, Mahdi and Busam, Benjamin and Ilic, Slobodan},
  journal={Advances in Neural Information Processing Systems},
  volume={34},
  pages={23872--23884},
  year={2021}
}

@inproceedings{lu2021hregnet,
  title={Hregnet: A hierarchical network for large-scale outdoor lidar point cloud registration},
  author={Lu, Fan and Chen, Guang and Liu, Yinlong and Zhang, Lijun and Qu, Sanqing and Liu, Shu and Gu, Rongqi},
  booktitle={Proceedings of the IEEE/CVF International Conference on Computer Vision},
  pages={16014--16023},
  year={2021}
}

@article{wang2022residual,
  title={Residual 3-d scene flow learning with context-aware feature extraction},
  author={Wang, Guangming and Hu, Yunzhe and Wu, Xinrui and Wang, Hesheng},
  journal={IEEE Transactions on Instrumentation and Measurement},
  volume={71},
  pages={1--9},
  year={2022},
  publisher={IEEE}
}

@article{sharp2002icp,
  title={ICP registration using invariant features},
  author={Sharp, Gregory C and Lee, Sang W and Wehe, David K},
  journal={IEEE Transactions on Pattern Analysis and Machine Intelligence},
  volume={24},
  number={1},
  pages={90--102},
  year={2002},
  publisher={IEEE}
}

@article{lin2022ds,
  title={Ds-transunet: Dual swin transformer u-net for medical image segmentation},
  author={Lin, Ailiang and Chen, Bingzhi and Xu, Jiayu and Zhang, Zheng and Lu, Guangming and Zhang, David},
  journal={IEEE Transactions on Instrumentation and Measurement},
  volume={71},
  pages={1--15},
  year={2022},
  publisher={IEEE}
}

@inproceedings{ghiasi2019fpn,
  title={Nas-fpn: Learning scalable feature pyramid architecture for object detection},
  author={Ghiasi, Golnaz and Lin, Tsung-Yi and Le, Quoc V},
  booktitle={Proceedings of the IEEE/CVF conference on computer vision and pattern recognition},
  pages={7036--7045},
  year={2019}
}

@article{xie2023cross,
  title={Cross-modal information-guided network using contrastive learning for point cloud registration},
  author={Xie, Yifan and Zhu, Jihua and Li, Shiqi and Shi, Pengcheng},
  journal={IEEE Robotics and Automation Letters},
  volume={9},
  number={1},
  pages={103--110},
  year={2023},
  publisher={IEEE}
}

@article{wang2022efficient,
  title={Efficient 3d deep lidar odometry},
  author={Wang, Guangming and Wu, Xinrui and Jiang, Shuyang and Liu, Zhe and Wang, Hesheng},
  journal={IEEE Transactions on Pattern Analysis and Machine Intelligence},
  volume={45},
  number={5},
  pages={5749--5765},
  year={2022},
  publisher={IEEE}
}

@inproceedings{ao2021spinnet,
  title={Spinnet: Learning a general surface descriptor for 3d point cloud registration},
  author={Ao, Sheng and Hu, Qingyong and Yang, Bo and Markham, Andrew and Guo, Yulan},
  booktitle={Proceedings of the IEEE/CVF conference on computer vision and pattern recognition},
  pages={11753--11762},
  year={2021}
}

@inproceedings{wang2019deep,
  title={Deep closest point: Learning representations for point cloud registration},
  author={Wang, Yue and Solomon, Justin M},
  booktitle={Proceedings of the IEEE/CVF international conference on computer vision},
  pages={3523--3532},
  year={2019}
}

@article{xu2024igreg,
  title={IGReg: Image-Geometry-Assisted Point Cloud Registration via Selective Correlation Fusion},
  author={Xu, Zongyi and Jiang, Xinqi and Gao, Xinyu and Gao, Rui and Gu, Changjun and Zhang, Qianni and Li, Weisheng and Gao, Xinbo},
  journal={IEEE Transactions on Multimedia},
  year={2024},
  publisher={IEEE}
}

@inproceedings{nie2022pyramid,
  title={Pyramid architecture for multi-scale processing in point cloud segmentation},
  author={Nie, Dong and Lan, Rui and Wang, Ling and Ren, Xiaofeng},
  booktitle={Proceedings of the IEEE/CVF conference on computer vision and pattern recognition},
  pages={17284--17294},
  year={2022}
}

@inproceedings{zhang20233d,
  title={3D registration with maximal cliques},
  author={Zhang, Xiyu and Yang, Jiaqi and Zhang, Shikun and Zhang, Yanning},
  booktitle={Proceedings of the IEEE/CVF conference on computer vision and pattern recognition},
  pages={17745--17754},
  year={2023}
}

@article{yang2015go,
  title={Go-ICP: A globally optimal solution to 3D ICP point-set registration},
  author={Yang, Jiaolong and Li, Hongdong and Campbell, Dylan and Jia, Yunde},
  journal={IEEE transactions on pattern analysis and machine intelligence},
  volume={38},
  number={11},
  pages={2241--2254},
  year={2015},
  publisher={IEEE}
}

@inproceedings{bai2021pointdsc,
  title={Pointdsc: Robust point cloud registration using deep spatial consistency},
  author={Bai, Xuyang and Luo, Zixin and Zhou, Lei and Chen, Hongkai and Li, Lei and Hu, Zeyu and Fu, Hongbo and Tai, Chiew-Lan},
  booktitle={Proceedings of the IEEE/CVF conference on computer vision and pattern recognition},
  pages={15859--15869},
  year={2021}
}

@inproceedings{qin2022geometric,
  title={Geometric transformer for fast and robust point cloud registration},
  author={Qin, Zheng and Yu, Hao and Wang, Changjian and Guo, Yulan and Peng, Yuxing and Xu, Kai},
  booktitle={Proceedings of the IEEE/CVF conference on computer vision and pattern recognition},
  pages={11143--11152},
  year={2022}
}

@article{xie2024hecpg,
  title={HECPG: hyperbolic embedding and confident patch-guided network for point cloud matching},
  author={Xie, Yifan and Zhu, Jihua and Li, Shiqi and Hu, Naiwen and Shi, Pengcheng},
  journal={IEEE Transactions on Geoscience and Remote Sensing},
  year={2024},
  publisher={IEEE}
}

@article{huang2022imfnet,
  title={IMFNet: Interpretable multimodal fusion for point cloud registration},
  author={Huang, Xiaoshui and Qu, Wentao and Zuo, Yifan and Fang, Yuming and Zhao, Xiaowei},
  journal={IEEE Robotics and Automation Letters},
  volume={7},
  number={4},
  pages={12323--12330},
  year={2022},
  publisher={IEEE}
}

@inproceedings{thomas2019kpconv,
  title={Kpconv: Flexible and deformable convolution for point clouds},
  author={Thomas, Hugues and Qi, Charles R and Deschaud, Jean-Emmanuel and Marcotegui, Beatriz and Goulette, Fran{\c{c}}ois and Guibas, Leonidas J},
  booktitle={Proceedings of the IEEE/CVF international conference on computer vision},
  pages={6411--6420},
  year={2019}
}

@inproceedings{bai2020d3feat,
  title={D3feat: Joint learning of dense detection and description of 3d local features},
  author={Bai, Xuyang and Luo, Zixin and Zhou, Lei and Fu, Hongbo and Quan, Long and Tai, Chiew-Lan},
  booktitle={Proceedings of the IEEE/CVF conference on computer vision and pattern recognition},
  pages={6359--6367},
  year={2020}
}

@inproceedings{chen2023sira,
  title={Sira-pcr: Sim-to-real adaptation for 3d point cloud registration},
  author={Chen, Suyi and Xu, Hao and Li, Ru and Liu, Guanghui and Fu, Chi-Wing and Liu, Shuaicheng},
  booktitle={Proceedings of the IEEE/CVF international conference on computer vision},
  pages={14394--14405},
  year={2023}
}

@inproceedings{zhang2022pcr,
  title={Pcr-cg: Point cloud registration via deep explicit color and geometry},
  author={Zhang, Yu and Yu, Junle and Huang, Xiaolin and Zhou, Wenhui and Hou, Ji},
  booktitle={European Conference on Computer Vision},
  pages={443--459},
  year={2022},
  organization={Springer}
}

@INPROCEEDINGS{9010002,
  author={Thomas, Hugues and Qi, Charles R. and Deschaud, Jean-Emmanuel and Marcotegui, Beatriz and Goulette, François and Guibas, Leonidas},
  booktitle={2019 IEEE/CVF International Conference on Computer Vision (ICCV)}, 
  title={KPConv: Flexible and Deformable Convolution for Point Clouds}, 
  year={2019},
  volume={},
  number={},
  pages={6410-6419},
  doi={10.1109/ICCV.2019.00651}}

@inproceedings{choy2019fully,
  title={Fully convolutional geometric features},
  author={Choy, Christopher and Park, Jaesik and Koltun, Vladlen},
  booktitle={Proceedings of the IEEE/CVF international conference on computer vision},
  pages={8958--8966},
  year={2019}
}

@inproceedings{choy2020deep,
  title={Deep global registration},
  author={Choy, Christopher and Dong, Wei and Koltun, Vladlen},
  booktitle={Proceedings of the IEEE/CVF conference on computer vision and pattern recognition},
  pages={2514--2523},
  year={2020}
}

@inproceedings{huang2020feature,
  title={Feature-metric registration: A fast semi-supervised approach for robust point cloud registration without correspondences},
  author={Huang, Xiaoshui and Mei, Guofeng and Zhang, Jian},
  booktitle={Proceedings of the IEEE/CVF conference on computer vision and pattern recognition},
  pages={11366--11374},
  year={2020}
}

@article{zhang2024adaptive,
  title={Adaptive decomposition and extraction network of individual fingerprint features for specific emitter identification},
  author={Zhang, Junning and Liu, Yicen and Ding, Guoru and Tang, Bo and Chen, Yanlong},
  journal={IEEE Transactions on Information Forensics and Security},
  year={2024},
  publisher={IEEE}
}

@inproceedings{jost2002fast,
  title={Fast ICP algorithms for shape registration},
  author={Jost, Timoth{\'e}e and H{\"u}gli, Heinz},
  booktitle={Pattern Recognition: 24th DAGM Symposium Zurich, Switzerland, September 16--18, 2002 Proceedings 24},
  pages={91--99},
  year={2002},
  organization={Springer}
}

@inproceedings{zhou2016fast,
  title={Fast global registration},
  author={Zhou, Qian-Yi and Park, Jaesik and Koltun, Vladlen},
  booktitle={Computer Vision--ECCV 2016: 14th European Conference, Amsterdam, The Netherlands, October 11-14, 2016, Proceedings, Part II 14},
  pages={766--782},
  year={2016},
  organization={Springer}
}

@article{zhang2023pyrf,
  title={Pyrf-pcr: A robust three-stage 3d point cloud registration for outdoor scene},
  author={Zhang, Junning and Huang, Siyuan and Liu, Jun and Zhu, Xiaoxiu and Xu, Feng},
  journal={IEEE Transactions on Intelligent Vehicles},
  volume={9},
  number={1},
  pages={1270--1281},
  year={2023},
  publisher={IEEE}
}

@inproceedings{gojcic2019perfect,
  title={The perfect match: 3d point cloud matching with smoothed densities},
  author={Gojcic, Zan and Zhou, Caifa and Wegner, Jan D and Wieser, Andreas},
  booktitle={Proceedings of the IEEE/CVF conference on computer vision and pattern recognition},
  pages={5545--5554},
  year={2019}
}

@article{fischler1981random,
  title={Random sample consensus: a paradigm for model fitting with applications to image analysis and automated cartography},
  author={Fischler, Martin A and Bolles, Robert C},
  journal={Communications of the ACM},
  volume={24},
  number={6},
  pages={381--395},
  year={1981},
  publisher={ACM New York, NY, USA}
}

@article{hu2024featsync,
  title={FeatSync: 3D point cloud multiview registration with attention feature-based refinement},
  author={Hu, Yiheng and Li, Binghao and Xu, Chengpei and Saydam, Sarp and Zhang, Wenjie},
  journal={Neurocomputing},
  volume={600},
  pages={128088},
  year={2024},
  publisher={Elsevier}
}

@inproceedings{huang2021predator,
  title={Predator: Registration of 3d point clouds with low overlap},
  author={Huang, Shengyu and Gojcic, Zan and Usvyatsov, Mikhail and Wieser, Andreas and Schindler, Konrad},
  booktitle={Proceedings of the IEEE/CVF Conference on computer vision and pattern recognition},
  pages={4267--4276},
  year={2021}
}

@inproceedings{li2020iterative,
  title={Iterative distance-aware similarity matrix convolution with mutual-supervised point elimination for efficient point cloud registration},
  author={Li, Jiahao and Zhang, Changhao and Xu, Ziyao and Zhou, Hangning and Zhang, Chi},
  booktitle={Computer Vision--ECCV 2020: 16th European Conference, Glasgow, UK, August 23--28, 2020, Proceedings, Part XXIV 16},
  pages={378--394},
  year={2020},
  organization={Springer}
}

@article{yuan2023egst,
  title={EGST: Enhanced geometric structure transformer for point cloud registration},
  author={Yuan, Yongzhe and Wu, Yue and Fan, Xiaolong and Gong, Maoguo and Ma, Wenping and Miao, Qiguang},
  journal={IEEE transactions on visualization and computer graphics},
  volume={30},
  number={9},
  pages={6222--6234},
  year={2023},
  publisher={IEEE}
}

@article{guo2024learning,
  title={Learning compact and overlap-biased interactions for point cloud registration},
  author={Guo, Lin and Chen, Zhi and Cheng, Senmao and Yang, Fan and Tao, Wenbing},
  journal={Neurocomputing},
  volume={598},
  pages={127949},
  year={2024},
  publisher={Elsevier}
}

@inproceedings{horache20213d,
  title={3D point cloud registration with multi-scale architecture and unsupervised transfer learning},
  author={Horache, Sofiane and Deschaud, Jean-Emmanuel and Goulette, Fran{\c{c}}ois},
  booktitle={2021 international conference on 3D vision (3DV)},
  pages={1351--1361},
  year={2021},
  organization={IEEE}
}

@inproceedings{yew2022regtr,
  title={Regtr: End-to-end point cloud correspondences with transformers},
  author={Yew, Zi Jian and Lee, Gim Hee},
  booktitle={Proceedings of the IEEE/CVF conference on computer vision and pattern recognition},
  pages={6677--6686},
  year={2022}
}

@article{huang2017coarse,
  title={A coarse-to-fine algorithm for matching and registration in 3D cross-source point clouds},
  author={Huang, Xiaoshui and Zhang, Jian and Wu, Qiang and Fan, Lixin and Yuan, Chun},
  journal={IEEE Transactions on Circuits and Systems for Video Technology},
  volume={28},
  number={10},
  pages={2965--2977},
  year={2017},
  publisher={IEEE}
}

@inproceedings{caesar2020nuscenes,
  title={nuscenes: A multimodal dataset for autonomous driving},
  author={Caesar, Holger and Bankiti, Varun and Lang, Alex H and Vora, Sourabh and Liong, Venice Erin and Xu, Qiang and Krishnan, Anush and Pan, Yu and Baldan, Giancarlo and Beijbom, Oscar},
  booktitle={Proceedings of the IEEE/CVF conference on computer vision and pattern recognition},
  pages={11621--11631},
  year={2020}
}

@article{an2024ol,
  title={OL-Reg: Registration of image and sparse LiDAR point cloud with object-level dense correspondences},
  author={An, Pei and Hu, Xuzhong and Ding, Junfeng and Zhang, Jun and Ma, Jie and Yang, You and Liu, Qiong},
  journal={IEEE Transactions on Circuits and Systems for Video Technology},
  year={2024},
  publisher={IEEE}
}

@inproceedings{lu2019deepvcp,
  title={Deepvcp: An end-to-end deep neural network for point cloud registration},
  author={Lu, Weixin and Wan, Guowei and Zhou, Yao and Fu, Xiangyu and Yuan, Pengfei and Song, Shiyu},
  booktitle={Proceedings of the IEEE/CVF international conference on computer vision},
  pages={12--21},
  year={2019}
}

@article{ren2022corri2p,
  title={CorrI2P: Deep image-to-point cloud registration via dense correspondence},
  author={Ren, Siyu and Zeng, Yiming and Hou, Junhui and Chen, Xiaodong},
  journal={IEEE Transactions on Circuits and Systems for Video Technology},
  volume={33},
  number={3},
  pages={1198--1208},
  year={2022},
  publisher={IEEE}
}

@inproceedings{wang2022improving,
  title={Improving rgb-d point cloud registration by learning multi-scale local linear transformation},
  author={Wang, Ziming and Huo, Xiaoliang and Chen, Zhenghao and Zhang, Jing and Sheng, Lu and Xu, Dong},
  booktitle={European Conference on Computer Vision},
  pages={175--191},
  year={2022},
  organization={Springer}
}

@inproceedings{wang2022you,
  title={You only hypothesize once: Point cloud registration with rotation-equivariant descriptors},
  author={Wang, Haiping and Liu, Yuan and Dong, Zhen and Wang, Wenping},
  booktitle={Proceedings of the 30th ACM International Conference on Multimedia},
  pages={1630--1641},
  year={2022}
}

@article{yang2022one,
  title={One-inlier is first: Towards efficient position encoding for point cloud registration},
  author={Yang, Fan and Guo, Lin and Chen, Zhi and Tao, Wenbing},
  journal={Advances in Neural Information Processing Systems},
  volume={35},
  pages={6982--6995},
  year={2022}
}

@article{wang2024neighborhood,
  title={Neighborhood multi-compound transformer for point cloud registration},
  author={Wang, Yong and Zhou, Pengbo and Geng, Guohua and An, Li and Li, Kang and Li, Ruoxue},
  journal={IEEE Transactions on Circuits and Systems for Video Technology},
  year={2024},
  publisher={IEEE}
}

@article{vizzo2023kiss,
  title={Kiss-icp: In defense of point-to-point icp--simple, accurate, and robust registration if done the right way},
  author={Vizzo, Ignacio and Guadagnino, Tiziano and Mersch, Benedikt and Wiesmann, Louis and Behley, Jens and Stachniss, Cyrill},
  journal={IEEE Robotics and Automation Letters},
  volume={8},
  number={2},
  pages={1029--1036},
  year={2023},
  publisher={IEEE}
}

@inproceedings{yuan2023pointmbf,
  title={PointMBF: A multi-scale bidirectional fusion network for unsupervised RGB-D point cloud registration},
  author={Yuan, Mingzhi and Fu, Kexue and Li, Zhihao and Meng, Yucong and Wang, Manning},
  booktitle={Proceedings of the IEEE/CVF International Conference on Computer Vision},
  pages={17694--17705},
  year={2023}
}

@inproceedings{yu2023rotation,
  title={Rotation-invariant transformer for point cloud matching},
  author={Yu, Hao and Qin, Zheng and Hou, Ji and Saleh, Mahdi and Li, Dongsheng and Busam, Benjamin and Ilic, Slobodan},
  booktitle={Proceedings of the IEEE/CVF conference on computer vision and pattern recognition},
  pages={5384--5393},
  year={2023}
}

@article{jiang2024gtinet,
  title={Gtinet: Global topology-aware interactions for unsupervised point cloud registration},
  author={Jiang, Yinuo and Zhou, Beitong and Liu, Xiaoyu and Li, Qingyi and Cheng, Cheng},
  journal={IEEE Transactions on Circuits and Systems for Video Technology},
  year={2024},
  publisher={IEEE}
}

@article{zhou2022sewer,
  title={Sewer defect detection from 3D point clouds using a transformer-based deep learning model},
  author={Zhou, Yunxiang and Ji, Ankang and Zhang, Limao},
  journal={Automation in construction},
  volume={136},
  pages={104163},
  year={2022},
  publisher={Elsevier}
}

@article{she2024pointdifformer,
  title={PointDifformer: Robust point cloud registration with neural diffusion and transformer},
  author={She, Rui and Kang, Qiyu and Wang, Sijie and Tay, Wee Peng and Zhao, Kai and Song, Yang and Geng, Tianyu and Xu, Yi and Navarro, Diego Navarro and Hartmannsgruber, Andreas},
  journal={IEEE Transactions on Geoscience and Remote Sensing},
  volume={62},
  pages={1--15},
  year={2024},
  publisher={IEEE}
}

@article{geiger2013vision,
  title={Vision meets robotics: The kitti dataset},
  author={Geiger, Andreas and Lenz, Philip and Stiller, Christoph and Urtasun, Raquel},
  journal={The international journal of robotics research},
  volume={32},
  number={11},
  pages={1231--1237},
  year={2013},
  publisher={Sage Publications Sage UK: London, England}
}

@article{sun2025hyperpoint,
  title={HyperPoint: Multimodal 3D foundation model in hyperbolic space},
  author={Sun, Yiding and Cheng, Haozhe and Lu, Chaoyi and Li, Zhengqiao and Wu, Minghong and Lu, Huimin and Zhu, Jihua},
  journal={Pattern Recognition},
  pages={112800},
  year={2025},
  publisher={Elsevier}
}

@inproceedings{han2025rethinking,
  title={Rethinking Regressor in 3D Gaussian Pretraining},
  author={Han, Xingguang and Sun, Yiding and Lu, Chaoyi},
  booktitle={Chinese Conference on Pattern Recognition and Computer Vision (PRCV)},
  pages={177--190},
  year={2025},
  organization={Springer}
}

@article{sun2026alignadaptrethinkingparameterefficient,
      title={Align then Adapt: Rethinking Parameter-Efficient Transfer Learning in 4D Perception}, 
      author={Yiding Sun and Jihua Zhu and Haozhe Cheng and Chaoyi Lu and Zhichuan Yang and Lin Chen and Yaonan Wang},
      journal={IEEE Trans. Multimedia},
      year={2026},
}

@article{sun6064487curve3d,
  title={Curve3D: Curvature-Aware Masked Autoencoder for Self-supervised Point Cloud Understanding},
  author={Sun, Yiding and Lu, Chaoyi and Cheng, Haozhe and Wang, Jun and Lu, Huimin and Chen, Lin and Zhu, Jihua},
  journal={Available at SSRN 6064487}
}

@article{wang2023enhancing,
  title={Enhancing power grid resilience with blockchain-enabled vehicle-to-vehicle energy trading in renewable energy integration},
  author={Wang, Yingsen and Zhang, Dongxu and Li, Yixiao and Jiao, Weihan and Wang, Guibin and Zhao, Juanjuan and Qiang, Yan and Li, Keqin},
  journal={IEEE Transactions on Industry Applications},
  volume={60},
  number={2},
  pages={2037--2052},
  year={2023},
  publisher={IEEE}
}

@article{li2023synergy,
  title={Synergy through integration of digital cognitive tests and wearable devices for mild cognitive impairment screening},
  author={Li, Aoyu and Li, Jingwen and Zhang, Dongxu and Wu, Wei and Zhao, Juanjuan and Qiang, Yan},
  journal={Frontiers in Human Neuroscience},
  volume={17},
  pages={1183457},
  year={2023},
  publisher={Frontiers Media SA}
}

@article{zhang2025ascot,
  title={Ascot: An adaptive self-correction chain-of-thought method for late-stage fragility in llms},
  author={Zhang, Dongxu and Yang, Ning and Zhu, Jihua and Yang, Jinnan and Xin, Miao and Tian, Baoliang},
  journal={arXiv preprint arXiv:2508.05282},
  year={2025}
}

@article{xue2024integrating,
  title={Integrating image and gene-data with a semi-supervised attention model for prediction of KRAS gene mutation status in non-small cell lung cancer},
  author={Xue, Yuting and Zhang, Dongxu and Jia, Liye and Yang, Wanting and Zhao, Juanjuan and Qiang, Yan and Wang, Long and Qiao, Ying and Yue, Huajie},
  journal={Plos one},
  volume={19},
  number={3},
  pages={e0297331},
  year={2024},
  publisher={Public Library of Science San Francisco, CA USA}
}

@article{zhang2026chain,
  title={Chain-of-Thought Compression Should Not Be Blind: V-Skip for Efficient Multimodal Reasoning via Dual-Path Anchoring},
  author={Zhang, Dongxu and Sun, Yiding and Tan, Cheng and Yan, Wenbiao and Yang, Ning and Zhu, Jihua and Zhang, Haijun},
  journal={arXiv preprint arXiv:2601.13879},
  year={2026}
}

@article{zhang2026pointcot,
  title={PointCoT: A Multi-modal Benchmark for Explicit 3D Geometric Reasoning},
  author={Zhang, Dongxu and Sun, Yiding and Li, Pengcheng and Liu, Yumou and Lin, Hongqiang and Xu, Haoran and Mu, Xiaoxuan and Lin, Liang and Yan, Wenbiao and Yang, Ning and others},
  journal={arXiv preprint arXiv:2602.23945},
  year={2026}
}

@article{zhang2026notallqueries,
      title={Not All Queries Need Deep Thought: CoFiCot for Adaptive Coarse-to-fine Stateful Refinement}, 
      author={Zhang, Dongxu and Lin, Hongqiang and Sun, Yiding and Wang, Pengyu and Wang, Qirui and Yang, Ning and Zhu, Jihua},
      journal={arXiv preprint arXiv:2603.08251},
      year={2026}
}

@article{gong2025med,
  title={Med-CMR: A Fine-Grained Benchmark Integrating Visual Evidence and Clinical Logic for Medical Complex Multimodal Reasoning},
  author={Gong, Haozhen and Ji, Xiaozhong and Liu, Yuansen and Wu, Wenbin and Yan, Xiaoxiao and Liu, Jingjing and Wu, Kai and Pan, Jiazhen and Jian, Bailiang and Zhang, Jiangning and others},
  journal={arXiv preprint arXiv:2512.00818},
  year={2025}
}

@article{lan2026reco,
  title={ReCo-KD: Region-and Context-Aware Knowledge Distillation for Efficient 3D Medical Image Segmentation},
  author={Lan, Qizhen and Hsu, Yu-Chun and Khan, Nida Saddaf and Jiang, Xiaoqian},
  journal={arXiv preprint arXiv:2601.08301},
  year={2026}
}

@article{chen2025uni,
  title={Uni-NTFM: A Unified Foundation Model for EEG Signal Representation Learning},
  author={Chen, Zhisheng and Zhang, Yingwei and Lan, Qizhen and Liu, Tianyu and Wang, Huacan and Ding, Yi and Jia, Ziyu and Chen, Ronghao and Wang, Kun and Zhou, Xinliang},
  journal={arXiv preprint arXiv:2509.24222},
  year={2025}
}

@article{lan2026performance,
  title={From Performance to Practice: Knowledge-Distilled Segmentator for On-Premises Clinical Workflows},
  author={Lan, Qizhen and Choi, Aaron and Ma, Jun and Wang, Bo and Zhao, Zhaogming and Jiang, Xiaoqian and Hsu, Yu-Chun},
  journal={arXiv preprint arXiv:2601.09191},
  year={2026}
}

@article{ZHANG2026133318,
title = {CMHANet: A cross-modal hybrid attention network for point cloud registration},
journal = {Neurocomputing},
pages = {133318},
year = {2026},
author = {Dongxu Zhang and Yingsen Wang and Yiding Sun and Haoran Xu and Peilin Fan and Jihua Zhu}
}

\end{document}